\pdfoutput=1

\documentclass[11pt]{article}

\usepackage{acl}

\usepackage{times}
\usepackage{latexsym}

\usepackage[T1]{fontenc}

\usepackage[utf8]{inputenc}

\usepackage{microtype}

\usepackage{inconsolata}

\usepackage{tabularx}
\usepackage{booktabs}
\usepackage{subcaption}
\usepackage{graphicx}
\usepackage{multirow}
\usepackage{enumitem}
\setlist{nosep}
\usepackage{soul}
\usepackage{pifont}
\usepackage{makecell}

\title{When is Tree Search Useful for LLM Planning?\\It Depends on the Discriminator}

\author{Ziru Chen$^1$, Michael White$^1$, Raymond Mooney$^2$, Ali Payani$^3$, Yu Su$^1$, Huan Sun$^1$ \\
  $^1$The Ohio State University \quad $^2$The University of Texas at Austin \quad $^3$Cisco Research\\
  \texttt{\{chen.8336, white.1240, su.809, sun.397\}@osu.edu} \\
  \texttt{mooney@cs.utexas.edu} \quad \texttt{apayani@cisco.com} \\
}

\begin{document}
\maketitle

\begin{abstract}
In this paper, we examine how large language models (LLMs) solve multi-step problems under a language agent framework with three components: a generator, a discriminator, and a planning method.
We investigate the practical utility of two advanced planning methods, iterative correction and tree search.
We present a comprehensive analysis of how discrimination accuracy affects the overall performance of agents when using these two methods or a simpler method, re-ranking.
Experiments on two tasks, text-to-SQL parsing and mathematical reasoning, show that: 
(1) advanced planning methods demand discriminators with at least 90\% accuracy to achieve significant improvements over re-ranking; 
(2) current LLMs' discrimination abilities have not met the needs of advanced planning methods to achieve such improvements; 
(3) with LLM-based discriminators, advanced planning methods may not adequately balance accuracy and efficiency. For example, compared to the other two methods, tree search is at least 10--20 times slower but leads to negligible performance gains, which hinders its real-world applications.\footnote{Code and data are available at \url{https://github.com/OSU-NLP-Group/llm-planning-eval}.}
\end{abstract}

\section{Introduction} 
\label{intro}

%
Planning plays a crucial role in intelligent behaviors of human and AI agents.
Since the early stage of AI research, various methods have been proposed to build agents that can plan efficiently and accurately \citep{newellsimon1956planning, russel2010}.
The problem-solving procedure in these AI agents usually involves three steps: searching for possible action sequences, predicting their expected outcomes with an internal world model, and finding an action sequence to achieve the best expected outcome \citep{russel2010, planning_brain}.
This procedure shares common traits with how large language models (LLMs) solve multi-step tasks, including mathematical reasoning \citep{wei2022cot}, multi-hop question answering \citep{yao2023react}, and code generation \citep{yang2023intercode}.
At each step, an LLM searches for possible next actions and generates their language representations (\textit{generation}).
To evaluate the actions, the LLM utilizes itself or another LLM to predict the outcomes of actions, in the form of rewards or correctness (\textit{discrimination}).
Afterwards, it incorporates the outcomes into its problem-solving process with some strategy to find the best action sequence (\textit{planning}).

\begin{figure}[t]
  \centering
  \includegraphics[width=0.85\columnwidth]{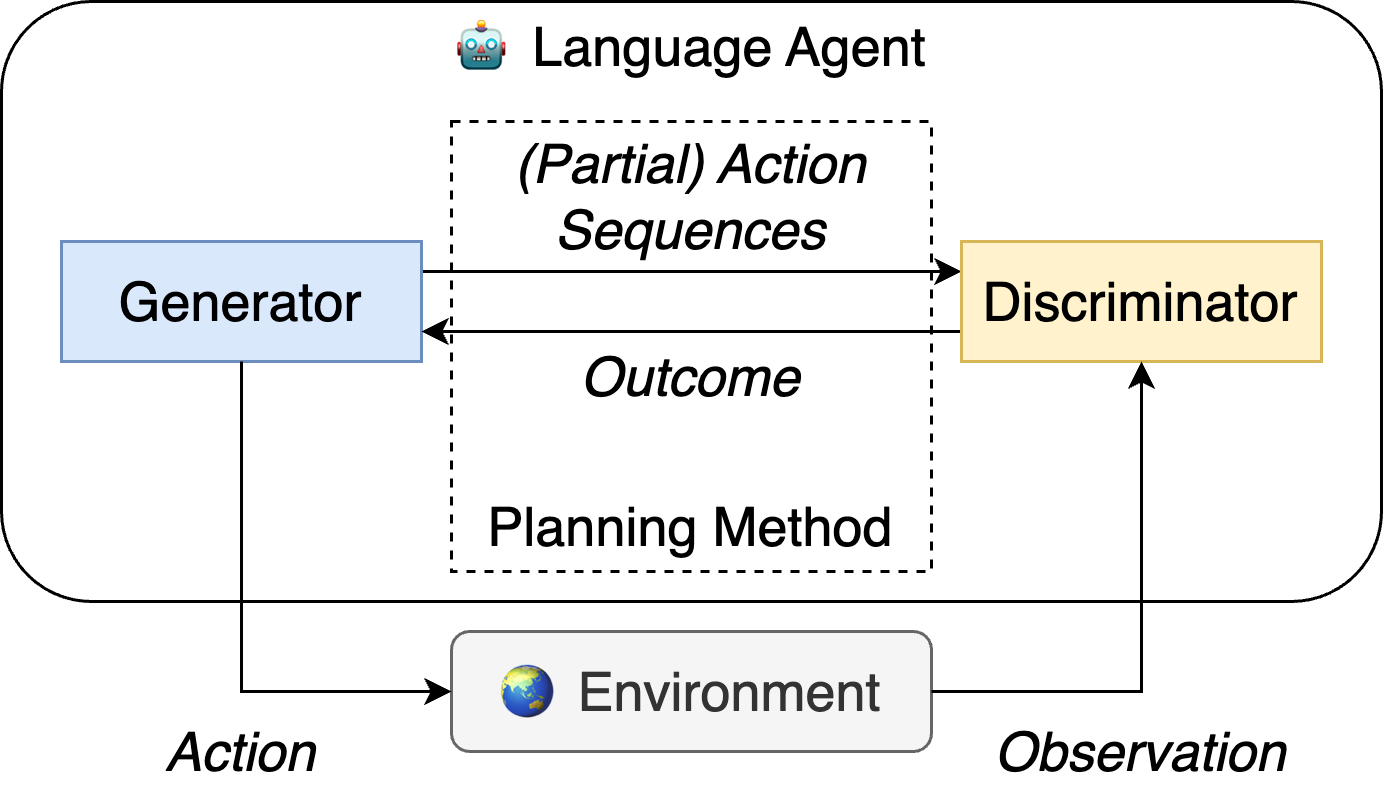}
  \vspace{-6pt}
  \caption{\label{fig:overview}A generator-discriminator framework of language agents, where planning methods control the interaction between a generator and a discriminator, both of which are usually instantiated by some LLM. 
  }
  \vspace{-16pt}
\end{figure}

Motivated by the similarity, we critically examine how LLMs solve multi-step tasks from a language-agent view.
We unify different problem-solving procedures of LLMs into an agent framework (Figure \ref{fig:overview}) consisting of a generator, a discriminator, and a planning method. 
Under this framework, we investigate the practical utility of more advanced planning methods, such as tree search, in comparison with simpler methods (e.g. re-ranking).
We hypothesize that the discriminator may be a deciding factor and systematically investigate two research questions:
\noindent \textbf{(RQ1)} \textit{How does discrimination accuracy affect the performance of language agents using different planning methods?}
\noindent \textbf{(RQ2)} \textit{Can LLM-based discriminators correctly assess language agents' actions in practical settings?}
To this end, we analyze LLMs' discrimination abilities and their impact on three categories of planning methods: re-ranking, iterative correction, and tree search. 
We comprehensively evaluate these methods on two real-world tasks, text-to-SQL parsing and mathematical reasoning, with open-source, proprietary, and fine-tuned LLM discriminators.
First, we use oracle environmental information to simulate discriminators with different levels of accuracy. 
The simulation experiments exhibit a strong correlation between discrimination accuracy and overall task performance among all three types of planning methods.
Then, in a non-oracle setting, we closely investigate the LLM-based discriminators and show how environmental observations can effectively improve them.
Finally, we conduct end-to-end evaluations of the discriminators and planning methods to verify and strengthen our findings.
In summary, our experiments show that:
\noindent \textbf{(1)} Advanced planning methods, i.e., iterative correction and tree search, demand highly accurate discriminators ($\ge 90\%$ accuracy) to achieve decent improvements over the simpler method, re-ranking.
\noindent \textbf{(2)} Using environmental feedback, we improve the discrimination accuracy of LLMs by up to 30.2 and 8.4 absolute points on text-to-SQL parsing and mathematical reasoning, respectively.
Yet, our end-to-end evaluations suggest they have barely met the need for advanced planning methods to show significant improvements over re-ranking.
%

%
\noindent \textbf{(3)} Meanwhile, advanced planning methods may not adequately balance accuracy and efficiency when using LLM-based discriminators.
In our experiments, compared to the other two methods, tree search is at least 10--20 times slower but leads to negligible performance gains. 
This accuracy-efficiency trade-off can impede the deployment of tree search in real-world applications.
%

\vspace{-2pt}
\section{Related Work}
\label{related_work_main}
\vspace{-2pt}

%
A lot of recent research efforts have focused on advanced planning methods for improving the multi-step problem-solving abilities of LLMs (\citealt{li-etal-2023-making, madaan2023selfrefine, yao2023tot, yao2023react, zhou2023language, feng2024alphazerolike}; \textit{inter alia}).
Despite different designs, all these methods use a discriminator to evaluate the agents' actions, or planning steps.
In fact, instead of planning methods, an agent's discriminator could be the more critical component.
Since incorrect outcome predictions could lead to suboptimal plans, discriminators may decide the performance of an agent, regardless of its planning method \citep{planning_brain}.
While it is commonly believed that discrimination is easier than generation for human and AI agents \citep{gu-etal-2023-dont}, \citet{west2024paradox} pose the hypothesis that state-of-the-art generative AI models, including LLMs, may not have discrimination abilities matching their generation abilities.
This hypothesis coincides with the findings of \citet{huang2024llm_selfcorrect} and \citet{wang-etal-2023-chatgpt-defend} that, without any external feedback or with obviously absurd feedback, LLMs may recognize some of their self-generated correct plans as wrong.
\citet{huang2024llm_selfcorrect} also note that the performance gains of self-correction, a kind of iterative correction method, may rely on some high-quality external feedback, such as checking ground-truth labels or test sets for planning loop termination.
However, such external feedback usually does not exist in practical applications because solutions to new problems are unknown, and annotating comprehensive test cases can be nontrivial and costly. 
Distinct from these existing studies, our work focuses on studying the relationship between discriminators and planning methods, including but not limited to self-correction, and attempts to improve LLMs' discrimination capability.
Our findings can provide useful guidelines for choosing planning methods and implementing language agents in practice.
In light of our findings, we encourage future research to thoroughly evaluate language agents with various practical, non-oracle discriminators.
We also advocate that improving LLM-based discriminators is an important future direction to enhance agents' accuracy and efficiency when using advanced planning methods.
%

\vspace{-2pt}
\section{Our Framework}
\label{framework}
\vspace{-2pt}

\begin{figure*}[t]
	\centering
	\footnotesize
	\begin{minipage}{0.32\linewidth}
		\centering
		\includegraphics[width=\linewidth]{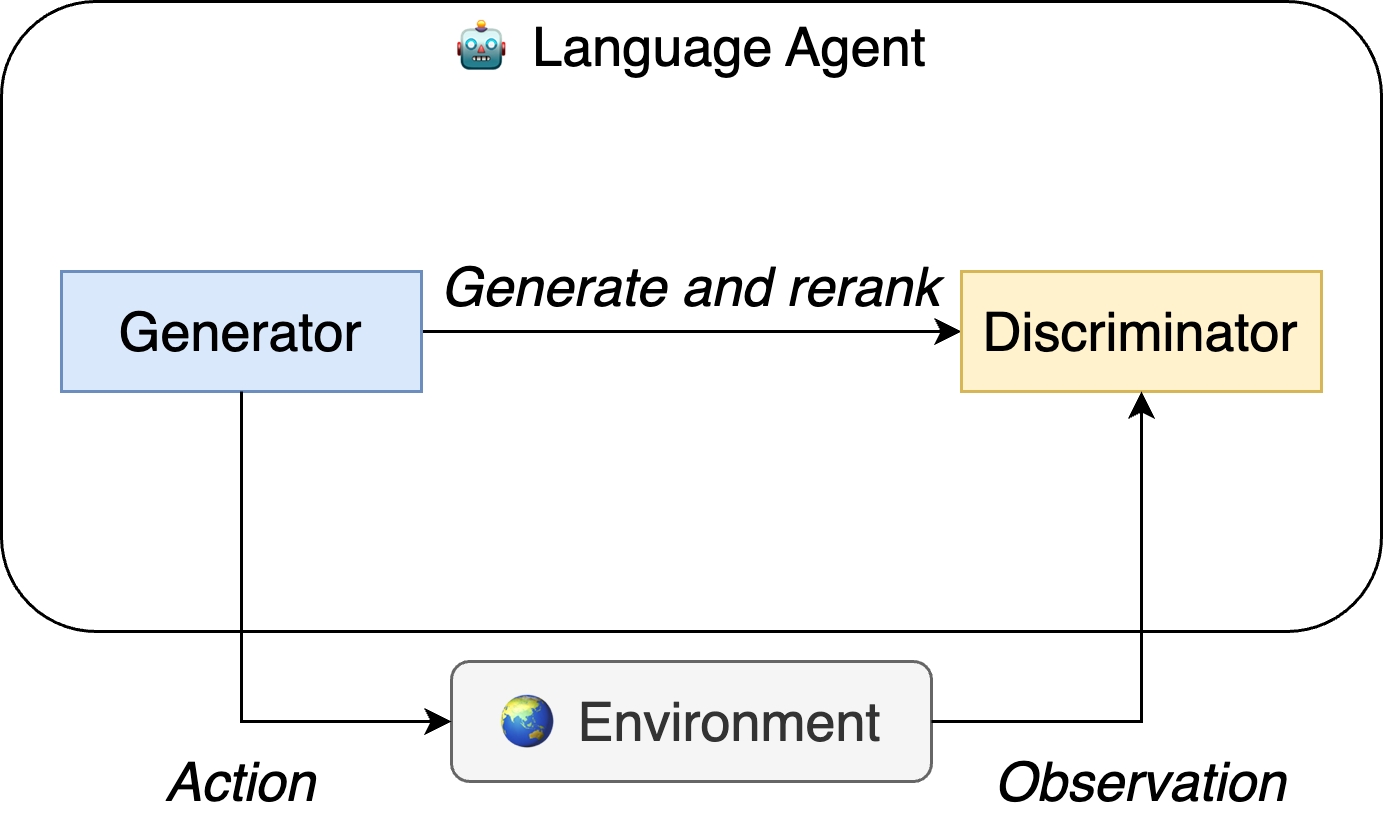}
		\subcaption{Re-ranking.}
		\label{fig:rr}
	\end{minipage}
        \hspace{0.01\linewidth}
	\begin{minipage}{0.32\linewidth}
		\centering
		\includegraphics[width=\linewidth]{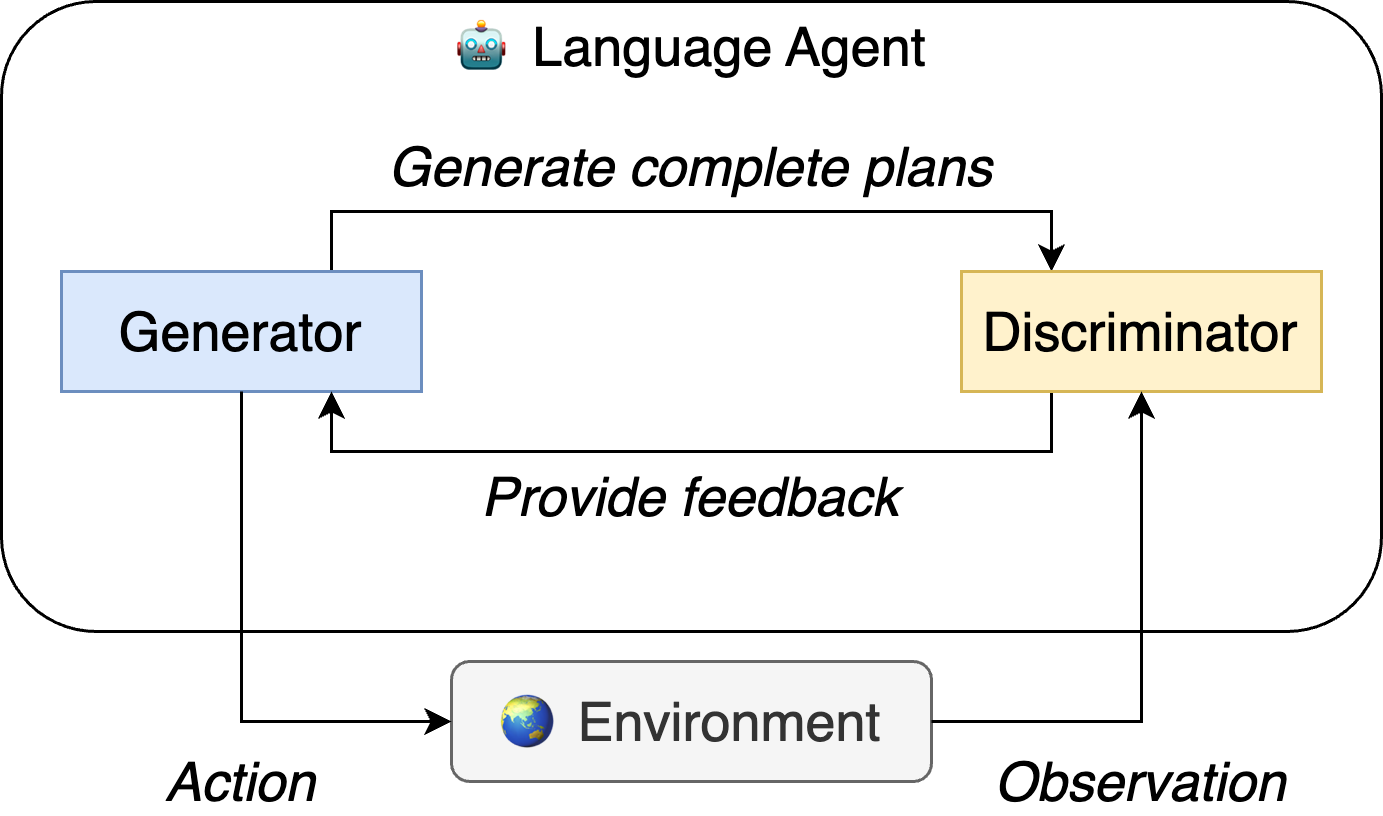}
		\subcaption{Iterative Correction.}
		\label{fig:ic}
	\end{minipage}
        \hspace{0.01\linewidth}
        \begin{minipage}{0.32\linewidth}
		\centering
		\includegraphics[width=\linewidth]{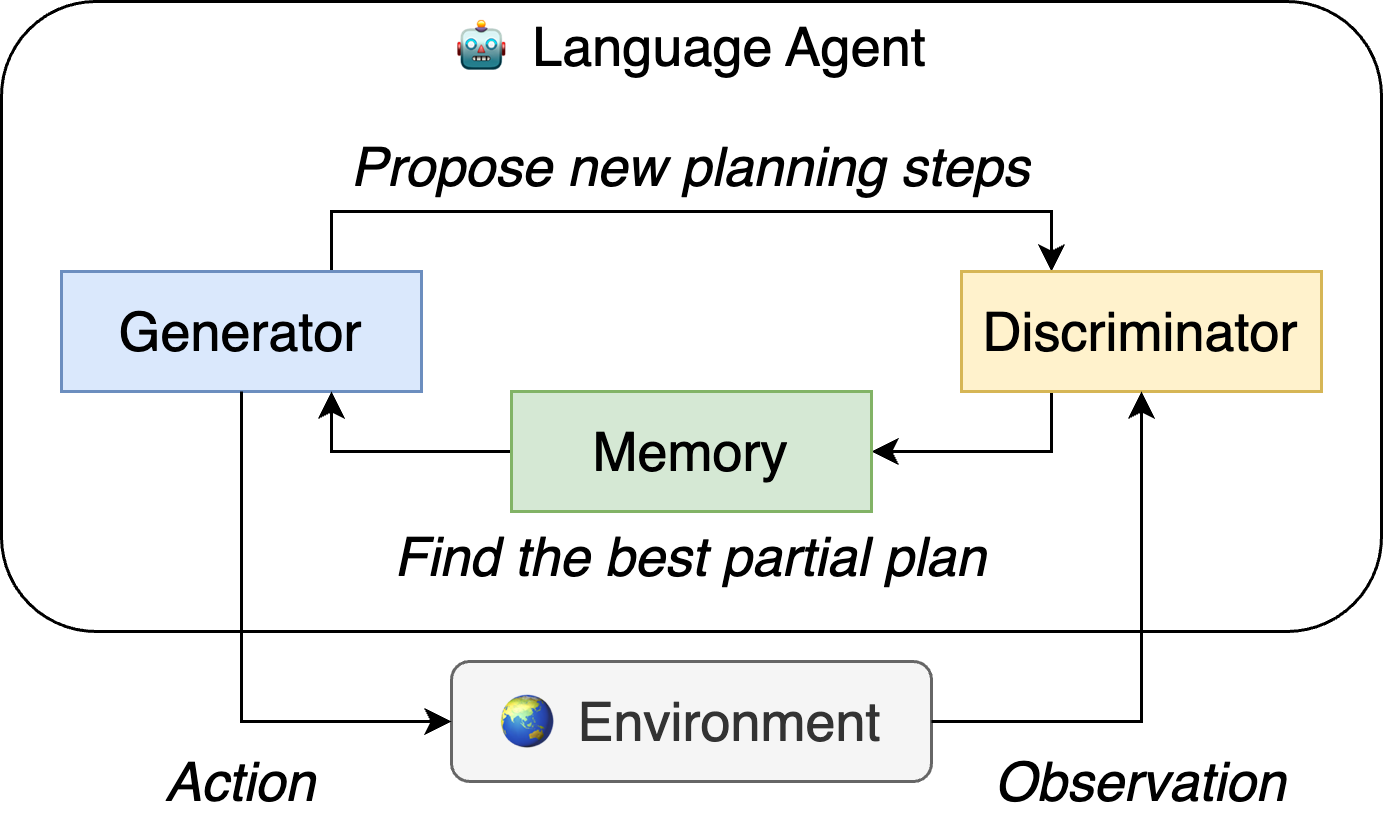}
		\subcaption{Tree Search.}
		\label{fig:ts}
	\end{minipage}
\vspace{-6pt}
\caption{Illustration of three categories of planning methods examined in our unified generator-evaluator framework. }
\vspace{-16pt}
\label{fig:methods}
\end{figure*}

As shown in Figure \ref{fig:overview}, we systematically analyze different planning methods in a unified generator-discriminator framework.
Our framework consists of a generator that proposes (partial) action sequences, a discriminator that evaluates the outcomes of these actions, and a planning method that ranks the actions according to their outcomes and manages the interaction between the two models.
In this section, we describe each of the three components and how they are instantiated on text-to-SQL parsing and mathematical reasoning (Section \ref{tasks_datasets}).
%

\subsection{Generator}
\label{generator}

%
For each planning step, we prompt the generator to sample action sequences (SQL queries or Python programs for math reasoning).
For text-to-SQL parsing, we use 1-shot prompting, where the example is retrieved from the training sets using BM25 \citep{bm25}.
For math reasoning, we use a fixed 2-shot prompt adapted from \citet{ni2023l2ceval}. 
See prompts in Appendix \ref{app:prompts}.
%

%
%

\subsection{Discriminator}
\label{discriminator}

%
Given some (partial) action sequences, we formulate the discrimination task as binary question answering \citep{kadavath2022language, ke-etal-2023-decompeval}. 
The discrimination score of each tested example is the probability of ``Yes'' being generated as the next token.
Specifically, we prompt the LLMs with the question ``Is the SQL/python program correct given the utterance/problem?'' to generate one single token with its probability as the score.
With this formulation, we evaluate three types of LLMs in our experiments (Section \ref{models}).
Similar to the generator, we use 1-shot prompting with BM25 retrieval for text-to-SQL parsing and a fixed 2-shot prompt for math reasoning.
Prompts are in Appendix \ref{app:prompts}.
%

\subsection{Planning Methods}
\label{planning_methods}

%
\noindent \textbf{Re-ranking.} 
Re-ranking is a straightforward planning method.
After sampling a few complete action sequences from the generator, it uses the discriminator to score them and return the highest-scoring plan (Figure \ref{fig:rr}).
Although simple, it is commonly used for code generation \citep{pmlr-v202-ni23b-lever} and mathematical reasoning tasks \citep{wang2023selfconsistency, li-etal-2023-making}.
We consider re-ranking as a baseline planning method for more advanced ones. 
\noindent \textbf{Iterative correction.} 
Like re-ranking, iterative correction starts with the generator proposing a complete action sequence.
Then it leverages multiple rounds of revision to improve the initial plan based on the discriminator's feedback (Figure \ref{fig:ic}).
When the generator and the discriminator are the same LLM, it becomes a prevalent planning method, self-correction \citep{madaan2023selfrefine, shinn2023reflexion, yao2023react, chen2024selfdebug}. 

While some work uses greedy generation, our implementation samples the same number of action sequences as other planning methods for fair comparison. 
Then, it uses the discriminator to select the best-scoring one for the next round's revision.
We allow up to 10 rounds of corrections, with early exiting when the best plan meets a threshold of discrimination score ($> 0.99$), or the score is not improved for 3 consecutive iterations.
For fair comparison, we prompt the generator to revise plans with 0-shot instruction following (Appendix \ref{app:prompts}) instead of few-shot, since in-context examples may introduce additional information.
\noindent \textbf{Tree Search.} 
Tree search is another popular planning method for language agents, such as Monte-Carlo Tree Search \citep{chaffin-etal-2022-ppl}, Pangu \citep{gu-etal-2023-dont}, RAP \citep{hao-etal-2023-reasoning}, Tree of Thoughts \citep{yao2023tot}, and LATS \citep{zhou2023language}. 
It uses a memory structure (e.g., a heap) to store observed partial action sequences and their scores.
For each iteration, it prompts the generator for possible next steps of the current best partial plan, calls the discriminator to evaluate the steps, and updates the memory with new plans and scores (Figure \ref{fig:ts}).
Our tree search implementation is a kind of MCTS \citep{zhang2023planning}:
\noindent \textbf{(1)} \textit{Selection}: Find the highest scoring partial plan in the memory, implemented as a heap structure.
\noindent \textbf{(2)} \textit{Expansion}: Prompt the generator for the next step of this partial plan. 
We follow recent work to define a step to be a SQL clause \citep{chen-etal-2023-text} or one line of Python code \citep{bui-etal-2022-detect}, which is semantically more meaningful. 
\noindent \textbf{(3)} \textit{Simulation}: Reuse the generator to complete the partial plans as Monte-Carlo simulations.
\noindent \textbf{(4)} \textit{Evaluation}: Evaluate the simulations with the discriminator.
The score for each new step is the maximum score of all simulations starting from it.
%

%
\noindent \textbf{(5)} \textit{Backpropagation}: Update the partial plan with the new step and score (if higher) and insert them into the heap memory.
After the update, if there is a complete plan in the heap memory, we terminate the tree search and return this plan.
%

\section{Experimental Setup}
\label{exp_setup}

\subsection{Tasks and Datasets}
\label{tasks_datasets}

%
\noindent \textbf{Text-to-SQL Parsing.} 
Text-to-SQL parsing is a code generation task of mapping natural language utterances to SQL queries. 
It requires agents to ground utterances to database environment and generate multi-step plans as SQL queries, making it an appropriate testbed in our study.
To evaluate language agents' potential for text-to-SQL parsing, we adapt two widely used datasets, Spider \citep{yu-etal-2018-spider} and Bird \citep{li2023bird}.
We use the entire training split in each dataset to prompt or fine-tune LLMs.\footnote{In Bird, we exclude training examples for one database, \texttt{retail\_world}, due to annotation errors.}
For evaluation, due to resource and budget constraints, we randomly select 400 and 300 development set examples in Spider and Bird, respectively.
We also note that model performance may be lower on our evaluation sets because we uniformly sampled examples from each difficulty level, while the original development sets have skewed distributions towards easier examples (Appendix \ref{app:text2sql_data}).
\noindent \textbf{Mathematical Reasoning.} 
Mathematical reasoning is a common task for evaluating language agents' multi-step reasoning and planning capabilities. 
With 500 random examples from GSM8K's development set \citep{cobbe2021gsm8k}, we follow program of thoughts \citep{chen2023pot} to test the agents' ability to plan in Python programs and solve these grade school math word problems.
%

\subsection{Models}
\label{models}
In all experiments, we use CodeLlama-13B-Instruct as the generator in our framework.
We also evaluate three kinds of LLMs as the discriminator:
\textbf{(1)} \textit{open-source LLMs}: CodeLlama-7B-Instruct and CodeLlama-13B-Instruct \citep{codellama},
\textbf{(2)} \textit{proprietary LLMs}: GPT-3.5-Turbo \citep{chatgpt} and GPT-4-Turbo \citep{openai2023gpt4}, and
\textbf{(3)} \textit{fine-tuned LLMs}: CodeLlama-7B-Instruct-FT and CodeLlama-13B-Instruct-FT.
For brevity, we will omit ``Instruct'' in model names.

\subsection{Implementation Details}
\label{implementation_details}
%
\noindent \textbf{Prompting the Generator LM.}
We prompt CodeLlama-13B with temperature-based sampling for different programs as action sequences (Appendix \ref{app:prompts}).
We use the model checkpoint and generation function implemented by HuggingFace \citep{wolf-etal-2020-transformers}.
We set the maximum generation length ($\mathtt{max\_length}$) to 300, temperature ($\mathtt{temperature}$) to 0.6, and number of samples ($\mathtt{num\_return\_sequences}$) to 5.
\noindent \textbf{Prompting Discriminator LMs.}
For CodeLlama-7B and CodeLlama-13B, we simply feed them the input prompt (Appendix \ref{app:prompts}) to get the last logit's values, which give us the token-level probability of ``Yes'' after applying the softmax function.
For GPT-3.5-Turbo (\texttt{gpt-3.5-turbo-1106}) and GPT-4-Turbo (\texttt{gpt-4-1106-preview}), we access them through the API of \citet{chatgpt, openai2023gpt4}.
We prompt the LLMs to generate one token and leverage the \texttt{top\_logprobs} request to check the top-5 tokens and their probabilities.\footnote{\url{https://platform.openai.com/docs/api-reference/chat/create\#chat-create-top\_logprobs}}
If ``Yes'' appears as one of the top-5 tokens, we take its probability $p$ without any modifications.
If ``Yes'' is missing and ``No'' appears as one of the top-5 tokens, we inverse its probability $1 - p$ as the score.
If both tokens are missing, our implementation returns 0, though this case should be rare in our experiments.
\noindent \textbf{Training Discriminator LMs.}
To get CodeLlama-7B-FT and CodeLlama-13B-FT, we again use the checkpoints and trainer implemented by HuggingFace.
We fine-tune the models with LoRA \citep{hu2022lora} to classify the correctness of a given program by generate one token: ``Yes'' or ``No''. Our training uses the following hyperparameters:
\begin{itemize}
    \item Number of epochs: 1
    \item Batch size: 128
    \item Learning rate: 1e-5
    \item Warmup ratio: 3\%
    \item Scheduler: cosine
\end{itemize}
The inference procedure of fine-tuned models is the same as how we prompt the pre-trained LLMs, but without using any in-context example.
\noindent \textbf{Computing Resources.} 
All of our experiments on Spider and GSM8K use up to four NVIDIA RTX A6000 GPU (48GB).
Experiments on Bird use up to four NVIDIA A100 Tensor Core GPU (80GB).

\begin{figure*}[t]
	\centering
	\footnotesize
	\begin{minipage}{0.33\linewidth}
		\centering
		\includegraphics[width=\linewidth]{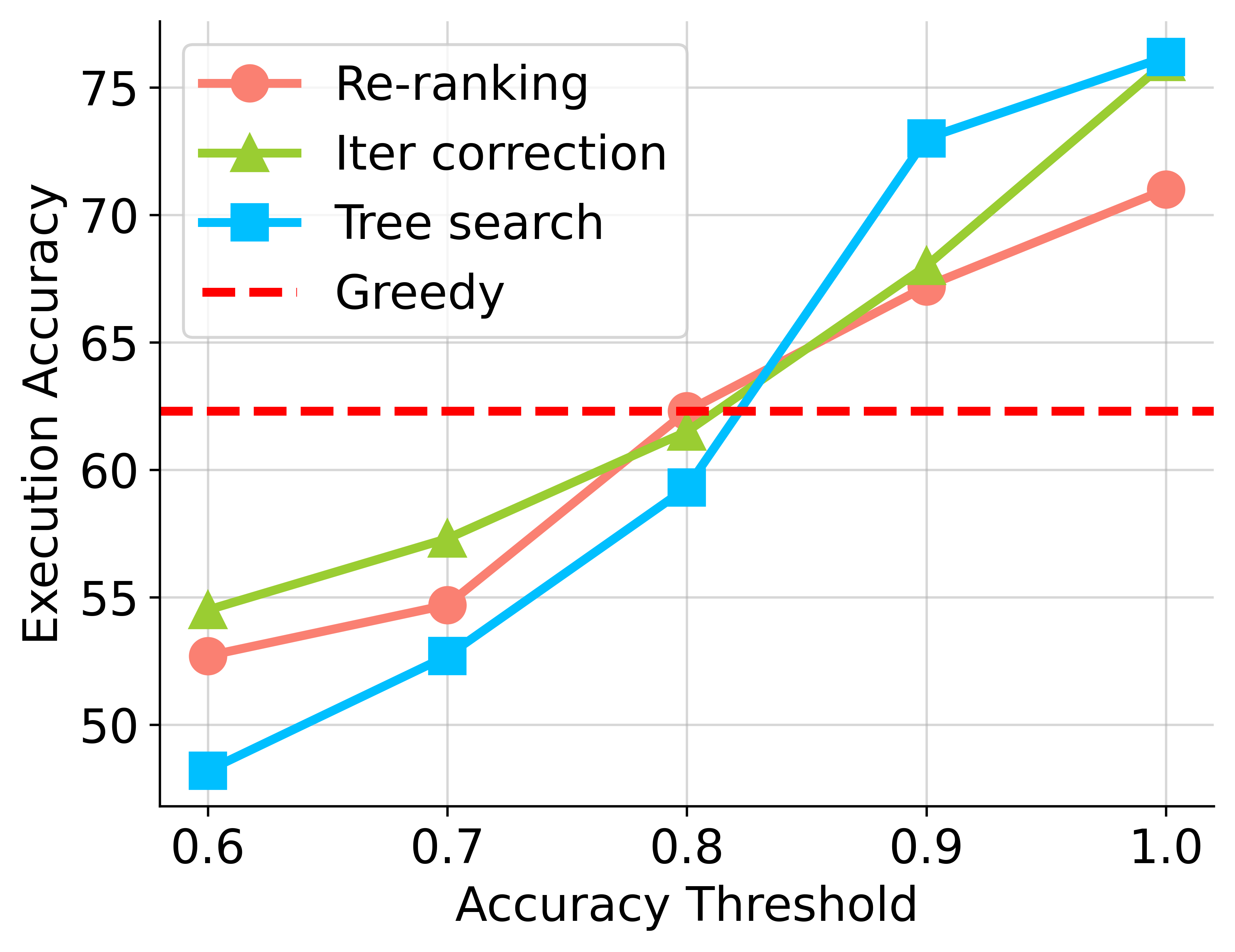}
	\end{minipage}
	%
        %
	\begin{minipage}{0.33\linewidth}
		\centering
		\includegraphics[width=\linewidth]{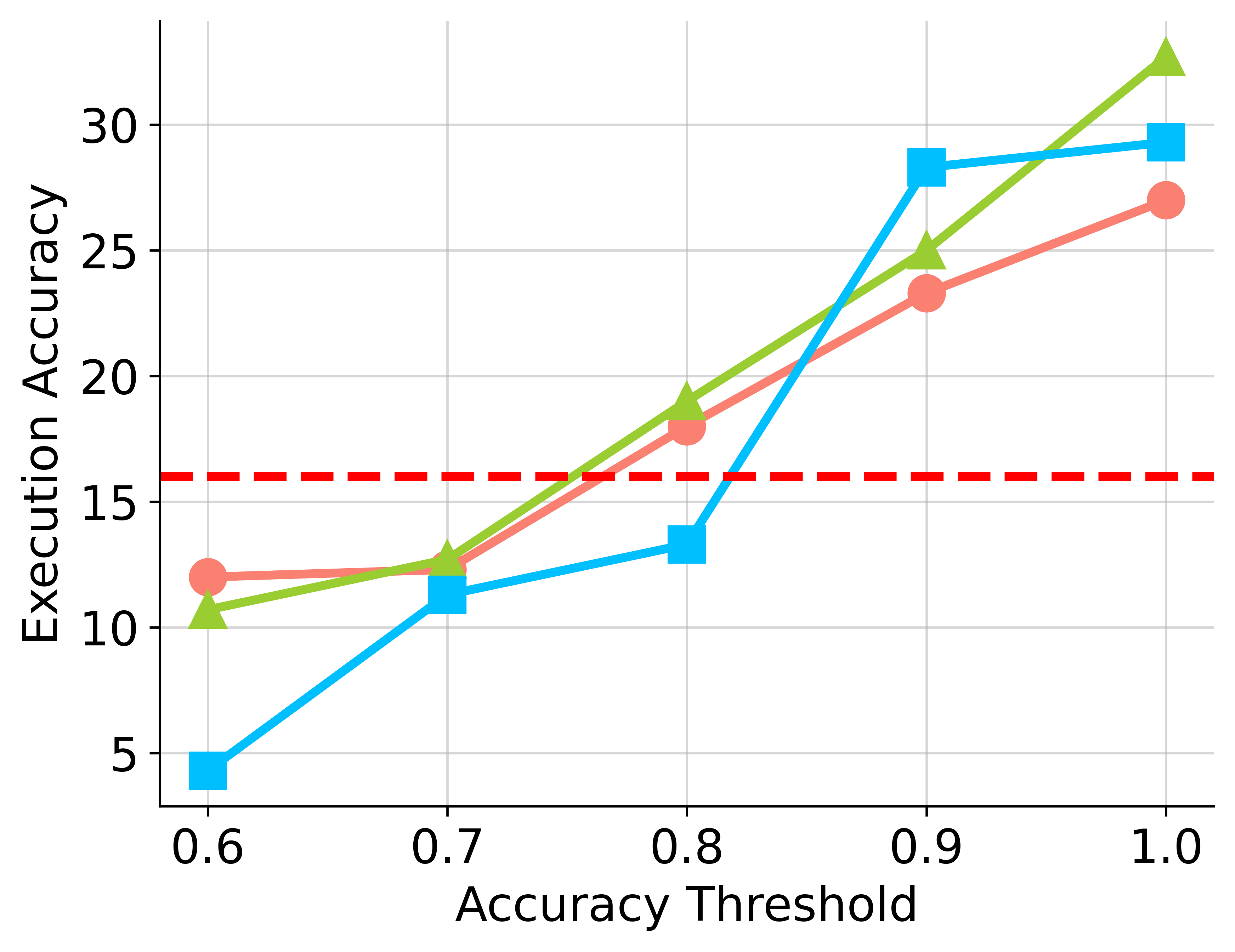}
	\end{minipage}
	%
        %
        \begin{minipage}{0.33\linewidth}
		\centering
		\includegraphics[width=\linewidth]{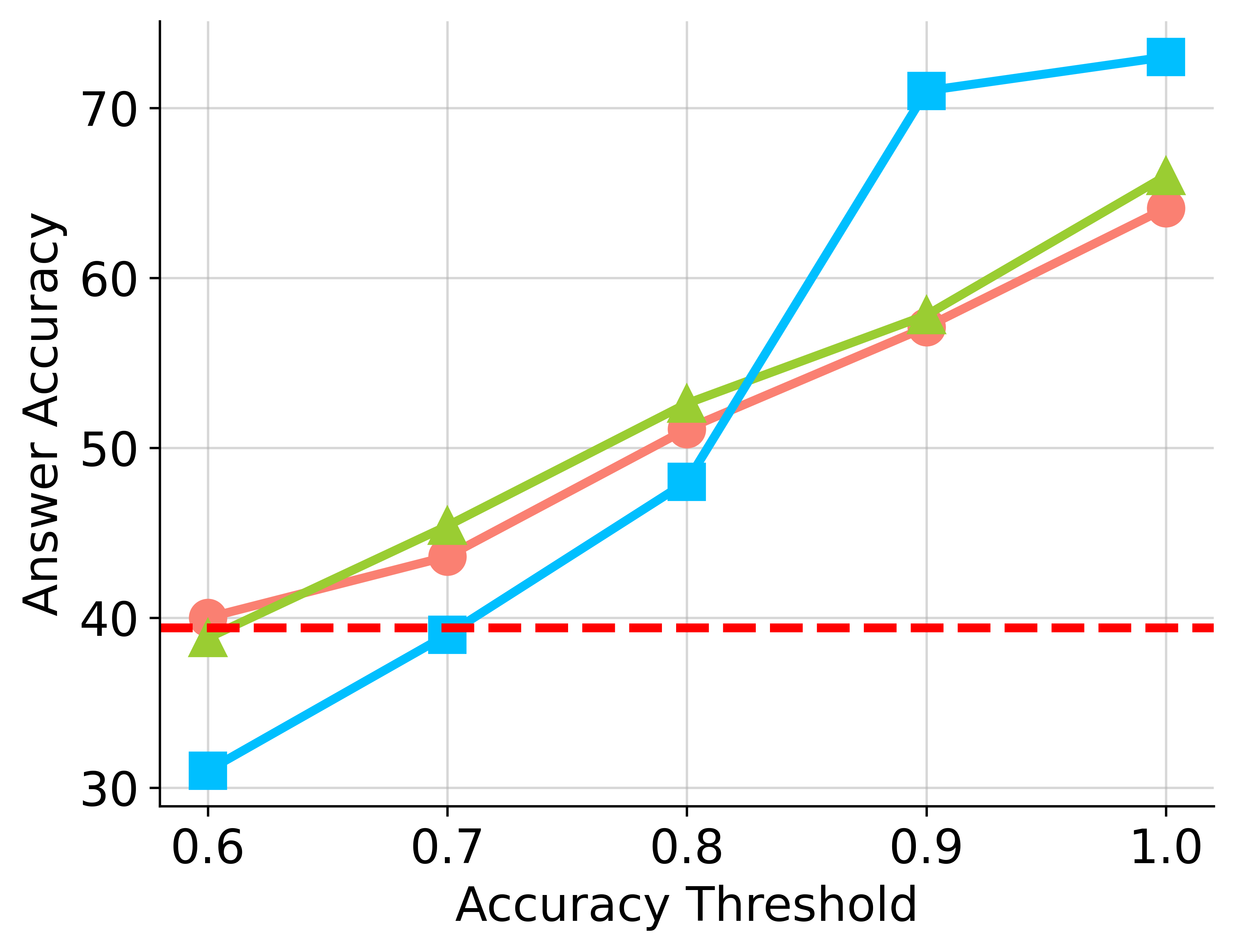}
	\end{minipage}

        \vspace{-11pt}
	\centering
	\footnotesize
	\begin{minipage}{0.33\linewidth}
		\centering
		\includegraphics[width=\linewidth]{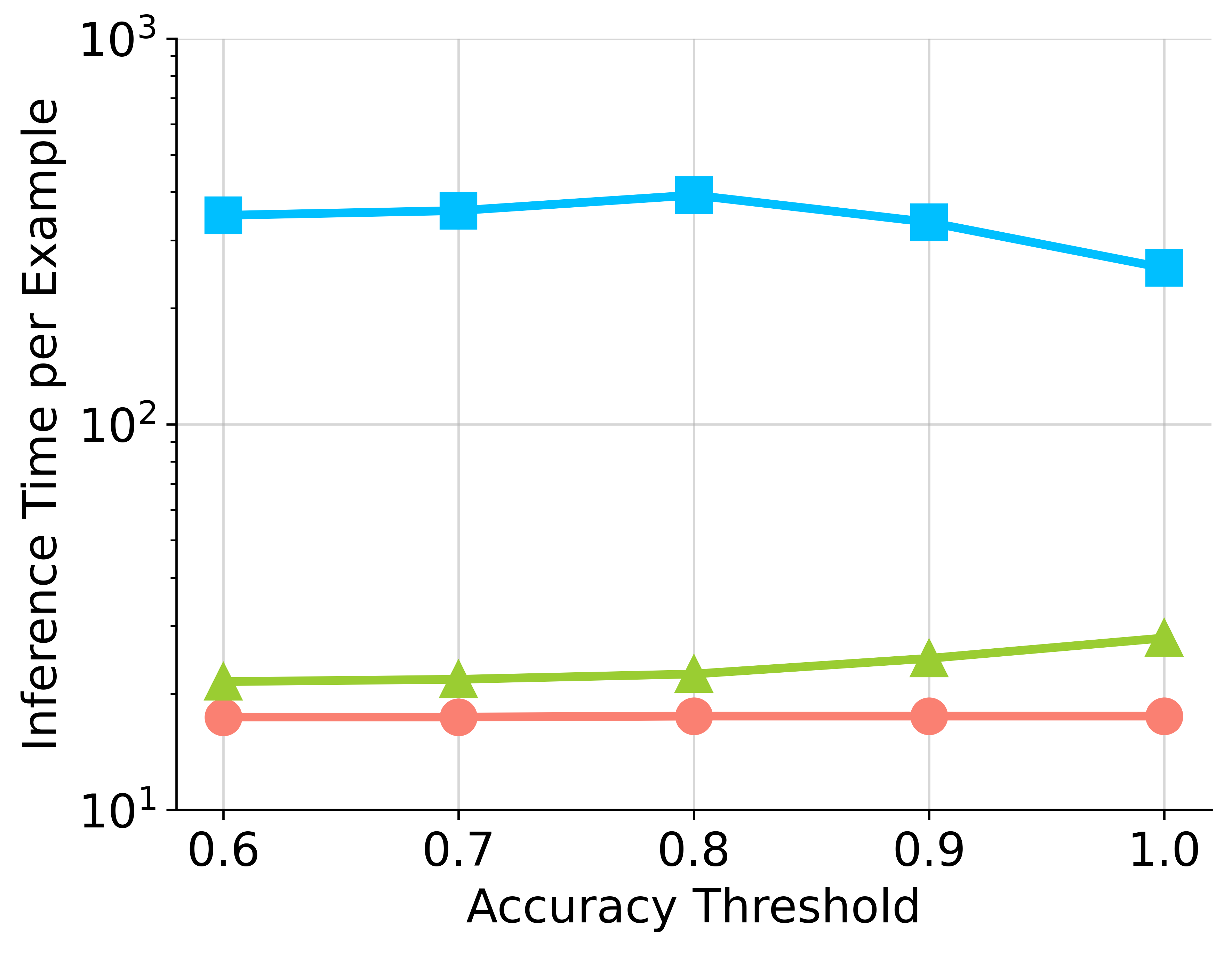}
		\subcaption{Spider.}
	\end{minipage}
	%
        %
	\begin{minipage}{0.33\linewidth}
		\centering
		\includegraphics[width=\linewidth]{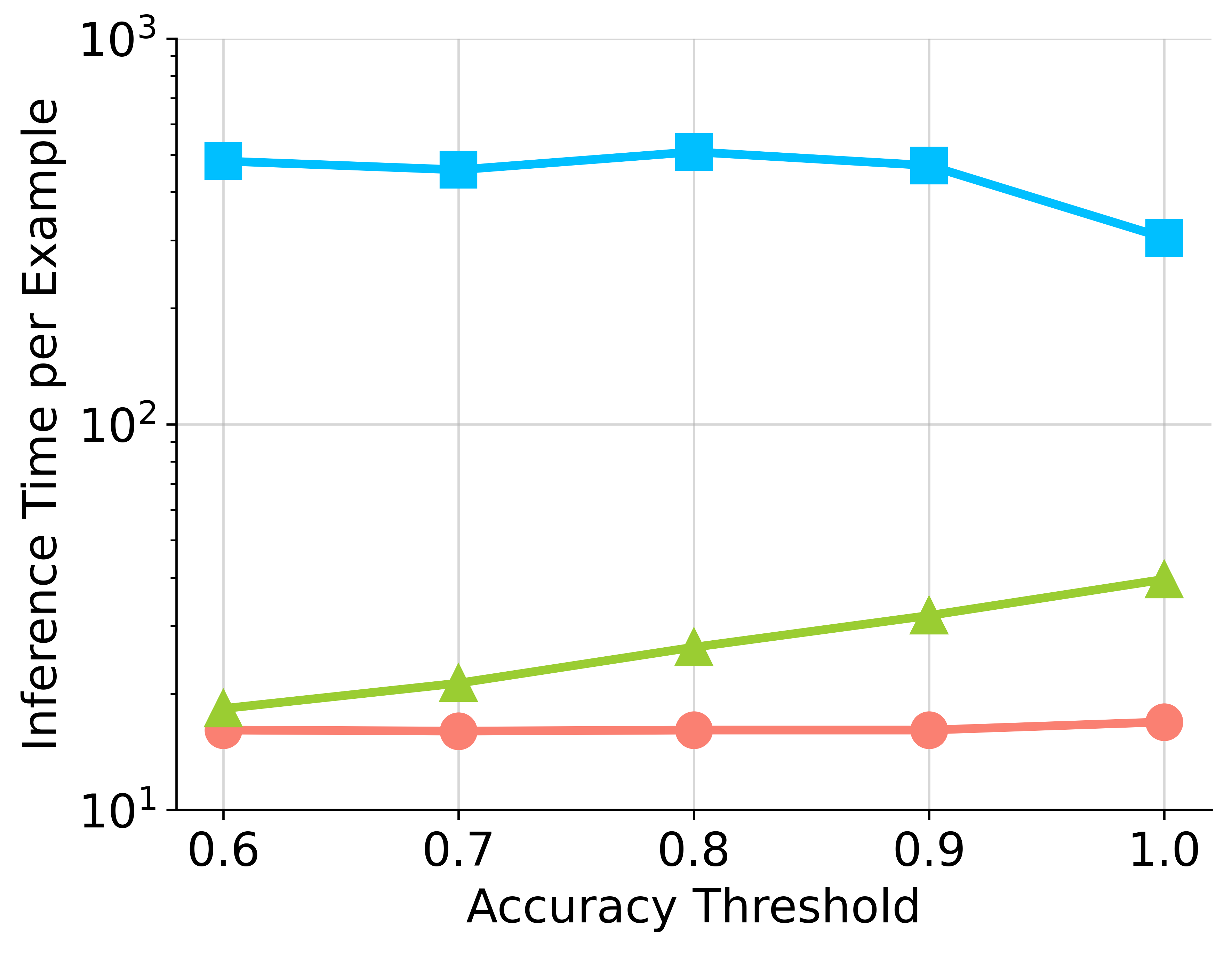}
		\subcaption{Bird.}
		\label{fig:bird}
	\end{minipage}
	%
        %
        \begin{minipage}{0.33\linewidth}
		\centering
		\includegraphics[width=\linewidth]{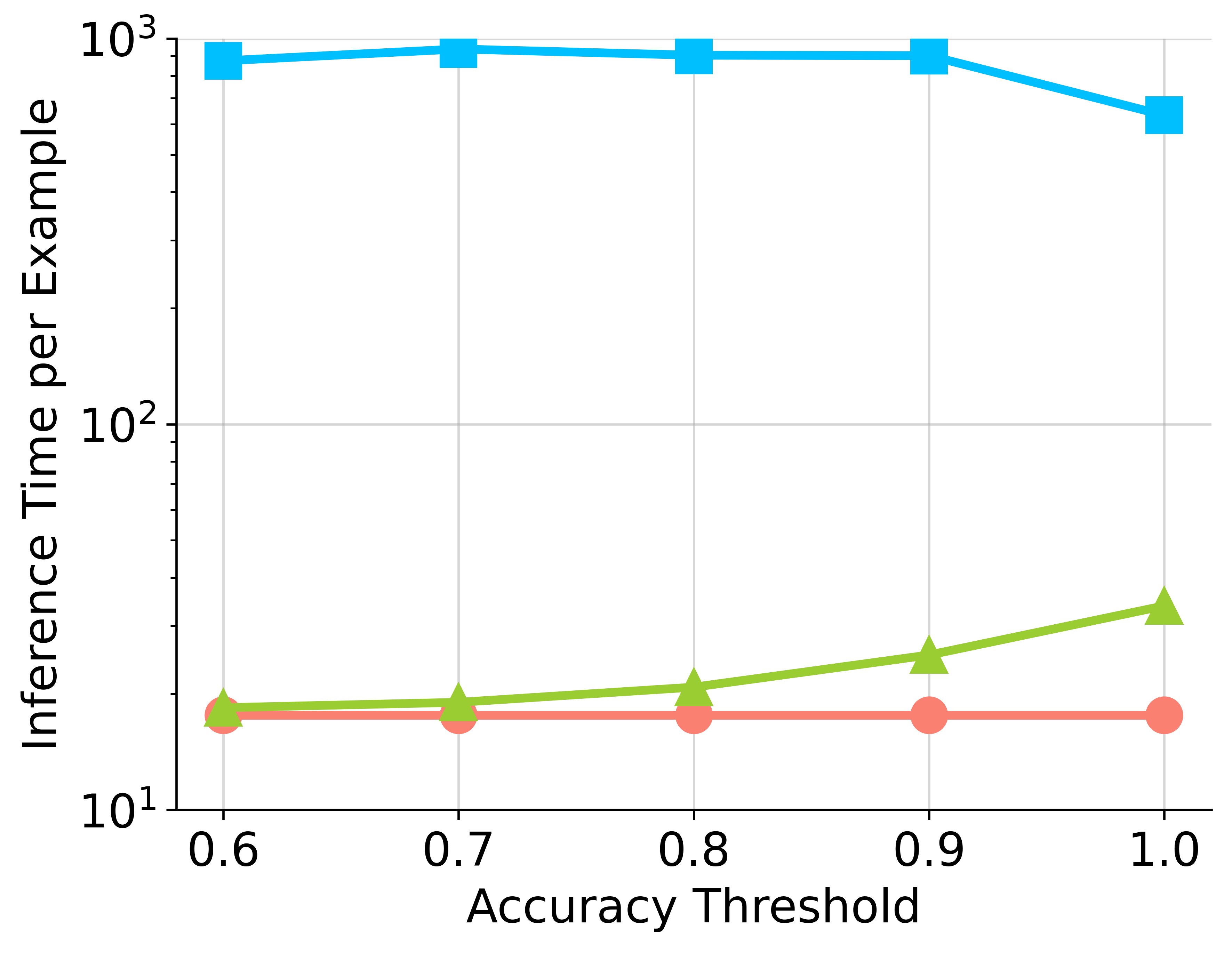}
		\subcaption{GSM8K.}
		\label{fig:gsm8k}
	\end{minipage}
\vspace{-6pt}
\caption{End-to-end evaluation results (the first row) and average inference time in log scale (the second row) of our simulation experiments with oracle. 
}
\vspace{-16pt}
\label{fig:oracle}
\end{figure*}

\subsection{Evaluation}
\label{eval}

%
\noindent \textbf{Intrinsic Evaluation.} 
We measure the discrimination abilities of LLMs with four intrinsic metrics.
\textbf{(1)} Discrimination accuracy (\textbf{Acc}): Given a pair of correct and wrong programs, we calculate the percentage where the correct program obtains a higher discrimination score than the wrong one \citep{bai2022training, touvron2023llama2}.
\textbf{(2)} Classification macro F1 (\textbf{F1}): We treat ``correct'' and ``wrong'' as two classes and compute the macro average of F1 scores on these two labels.
\textbf{(3)} Hit@1 (\textbf{H@1}): Given a batch of candidate programs, we calculate the percentage where the highest scoring candidate is correct.
\textbf{(4)} Mean reciprocal rank (\textbf{MRR}): We compute the standard MRR score by the highest-ranking correct program in the batches.
\noindent \textbf{End-to-End Evaluation.} 
To show the impact of discriminators, we evaluate language agents' end-to-end performance using our three planning methods, with execution accuracy for text-to-SQL parsing and answer accuracy for math reasoning.
%

\section{Simulation Experiments with Oracle}
\label{simulated_exp}

\subsection{Oracle-Based Discriminator}
\label{oracle}

%
To investigate how discrimination accuracy affects the overall performance of language agents using different planning methods (\textit{RQ1}), we utilize oracle environmental feedback to simulate a discriminator with controllable accuracy.
For text-to-SQL parsing, we compare the first five rows in the execution results of predicted and gold SQL queries and calculate their table cell overlaps (Appendix \ref{app:oracle_implementation}). 
For mathematical reasoning, we compare the predicted Python programs' answers with the ground truth.
We use a probability-based threshold $\tau$ to control the accuracy of each simulated discriminator \citep{gao-etal-2022-simulating}.
When evaluating each plan, the discriminator first computes a score $s$ with oracle information.
Then, it uses a random function to generate a number $p \in [0, 1)$.
If $p < \tau$, the discriminator returns the score $s$.
Otherwise, it returns an inverted score $1 - s$.
In this way, we ensure that the discriminator's accuracy is at most $\tau$.

\begin{table*}[t]
\small
\centering
\begin{tabular}{lcccccccccccc}
\toprule
\multirow{2}{*}{\textbf{Models}} & \multicolumn{4}{c}{\textbf{Spider}} & \multicolumn{4}{c}{\textbf{Bird}} & \multicolumn{4}{c}{\textbf{GSM8K}$^\ddag$} \\
\cmidrule(lr){2-5}\cmidrule(lr){6-9}\cmidrule(lr){10-13}
 & Acc & F1 & H@1 & MRR & Acc & F1 & H@1 & MRR & Acc & F1 & H@1 & MRR \\
\midrule
CodeLlama-7B & 54.0 & 37.1 & 56.0 & 62.3 & 44.6 & \textbf{46.7} & 13.0 & 18.0 & 48.6 & \textbf{38.7} & 36.2 & 46.9 \\
CodeLlama-13B & 58.2 & 37.1 & 57.0 & 63.1 & 49.4 & \textbf{46.7} & 12.7 & 18.3 & \textbf{62.2} & \textbf{38.7} & \textbf{41.8} & \textbf{51.0} \\
CodeLlama-7B-FT & 62.4 & 60.3 & 59.5 & 64.6 & 52.4 & \textbf{46.7} & 14.3 & 19.1 & - & - & - & - \\
CodeLlama-13B-FT & \textbf{69.7} & \textbf{67.2} & \textbf{61.3} & \textbf{65.7} & \textbf{62.1} & \textbf{46.7} & \textbf{16.0} & \textbf{20.5} & - & - & - & - \\
\midrule
GPT-3.5-Turbo & 67.0 & 47.3 & 59.0 & 64.3 & 64.3 & 35.7 & 16.0 & 20.5 & 72.1 & 49.1 & 46.6 & 54.0 \\
GPT-4-Turbo & \textbf{76.5} & \textbf{54.9} & \textbf{63.0} & \textbf{66.7} & \textbf{76.2} & \textbf{50.1} & \textbf{20.3} & \textbf{23.0} & \textbf{93.8} & \textbf{91.1} & \textbf{59.8} & \textbf{61.6} \\
\bottomrule
\end{tabular}
\vspace{-6pt}
\caption{\label{tab:intrin_eval}
Intrinsic evaluation results of naive LLMs' discrimination abilities. The \textbf{best performance} is in bold for open-source and closed-source LLMs. $^\ddag$Since GSM8K's training set does not have program of thoughts annotated for fine-tuning, we have only evaluated the models with in-context learning. 
}
\vspace{-11pt}
\end{table*}

\subsection{Results and Analysis}
\label{oracle_results}

%
As shown in Figure \ref{fig:oracle}, discrimination accuracy closely correlates with the performance of agents on all three datasets, no matter which planning method is used.
For instance, the performance of re-ranking agents improves linearly as we increase the discrimination accuracy threshold, setting up a strong baseline for agents using other planning methods.
We also note that it takes around $80\%$ discrimination accuracy for all agents to outperform greedy generation on text-to-SQL parsing, demonstrating the task's difficulty.
To answer \textit{RQ1}, we further analyze the performance of agents using iterative correction and tree search as follows:

\textbf{Advanced planning methods demand highly accurate discriminators.}
For iterative correction agents, their performance usually cannot distinguish from the re-ranking baselines until we maximize the threshold $\tau = 1.0$ (Figure \ref{fig:oracle}).
This finding resonates with \citet{huang2024llm_selfcorrect} that high-quality feedback may be the key to the success of iterative correction.
More interestingly, tree search agents consistently underperform the other two when the discrimination accuracy threshold $\tau \le 0.8$. 
Moreover, when raising the threshold to $0.9$, we observe a sharp increase of their performance, with which they start to beat other kinds of agents.
\textbf{Advanced planning methods may not adequately balance accuracy and efficiency.}
By calculating the average inference time per example (Figure \ref{fig:oracle}), we find that our implementation of tree search is at least $10 $--$20$ times slower than the other two planning methods, mainly due to frequent generation of Monte-Carlo simulations \citep{zhang2023planning}. 
While we can remove the simulations to be more efficient and evaluate partial plans, in our preliminary study, we find LLMs would struggle in this setting. 
This accuracy-efficiency trade-off may hinder real-world applications of tree search methods.
Meanwhile, the inference time for iterative correction increases as the accuracy threshold is raised, suggesting more iterations are required to derive a correct answer (Appendix \ref{app:ic_explanation}).
This indicates that developing efficient and accurate planning methods remains a key problem for AI agents.
\textbf{Monte-Carlo tree search can be unstable, especially in the early stages.}
We observe that iterative correction outperforms tree search on Bird (Figure \ref{fig:bird}) when the accuracy threshold is 1.0.
This observation may be caused by the instability of Monte-Carlo tree search.
We first note that McNemar’s test finds no difference between iterative correction and tree search ($p > 0.05$), despite their performance gap (29.3 vs 32.7). 
The rationales are discussed in Appendix \ref{app:stat_test}.
Furthermore, we analyze all 25 examples of which iterative correction derives the correct answer but tree search fails.
In 12 out of the 25 examples (48\%), tree search fails to select the correct partial plan when the discrimination scores are the same.
Especially, this can happen in the early stages of tree search, where a correct program has not yet been discovered and all the steps receive a score of 0 from the oracle discriminator.
Thus, we consider this underperformance a consequence of search instability.
%

\begin{table*}[t]
\small
\centering
\begin{tabular}{lcccccccc}
\toprule
 & \multicolumn{3}{c}{\textbf{CodeLlama-13B}} & \multicolumn{3}{c}{\textbf{GPT-3.5-Turbo}} & \multicolumn{2}{c}{\textbf{CodeLlama-13B-FT}} \\
\cmidrule(lr){2-4}\cmidrule(lr){5-7}\cmidrule(lr){8-9}
 & Spider & Bird & GSM8K & Spider & Bird & GSM8K & \hspace{4pt} Spider \hspace{4pt} & \hspace{4pt} Bird \hspace{4pt} \\
\midrule
Naive Discriminator & 58.2 & 49.4 & 62.2 & 67.0 & 64.3 & 72.1 & 69.7 & 62.1 \\
\midrule
+ Executability Check & 78.7 & 78.8 & 64.5 & 84.8 & 86.3 & 73.2 & 83.6 & 82.2 \\
++ Execution Result & \textbf{83.6} & \textbf{79.6} & \textbf{70.6} & \textbf{90.0} & \textbf{89.2} & \textbf{76.5} & \textbf{88.5} & \textbf{85.1} \\
\midrule
Improvement & \underline{25.4} & \underline{30.2} & \underline{8.4} & 23.0 & 24.9 & 4.4 & 18.8 & 23.0 \\
\bottomrule
\end{tabular}
\vspace{-6pt}
\caption{\label{tab:enhanced_llm}
Discrimination accuracy of observation-enhanced LLMs. The \textbf{best performance} (in bold) is achieved using both kinds of environmental observations. We also underline the \underline{largest improvement} for each dataset.
}
\vspace{-6pt}
\end{table*}
\begin{table*}[t]
\small
\centering
\begin{tabular}{lcccccc}
\toprule
\multirow{2}{*}{\textbf{Discriminators}} & \multicolumn{3}{c}{\textbf{Spider (Greedy Gen = 62.3)}} & \multicolumn{3}{c}{\textbf{Bird (Greedy Gen = 16.0)}} \\
\cmidrule(lr){2-4}\cmidrule(lr){5-7}
 & Re-ranking & Iter. Correct. & Tree Search & Re-ranking & Iter. Correct. & Tree Search \\
\midrule
CodeLlama-13B & \textbf{57.5} & 51.7 & 55.5 & \textbf{13.3} & \textbf{13.3} & \textbf{13.3} \\
GPT-3.5-Turbo & \textbf{58.3} & 52.7 & 56.2 & \underline{\textbf{18.0}} & 17.3 & 14.0 \\
CodeLlama-13B-FT & \underline{\textbf{61.5}} & 51.7 & 56.0 & \textbf{14.3} & 13.0 & 13.0 \\
\midrule
CodeLlama-13B$^{E}$ & \textbf{65.5} & 62.0 & 62.5 & 21.0 & \textbf{24.3} & 22.7 \\
GPT-3.5-Turbo$^{E}$ & 67.0 & \textbf{67.5} & 66.0 & 22.3 & \textbf{25.0} & 22.7 \\
CodeLlama-13B-FT$^{E}$ & \underline{\textbf{70.3}} & 68.0 & 67.5 & 23.7 & \underline{\textbf{26.3}} & 21.7 \\
\midrule
Oracle Simulation ($\tau = 1.0$) & 71.0 & 76.0$^*$ & 76.2$^*$ & 27.0 & 32.7$^*$ & 29.3 \\
\bottomrule
\end{tabular}
\vspace{-6pt}
\caption{\label{tab:text2sql}
End-to-end execution accuracy on text-to-SQL parsing. The \textbf{best performance} for each discriminator is in bold. The \underline{overall best performance} for naive and enhanced discriminators on each dataset is underlined. $^E$Observation-enhanced discriminators. $^*$Statistically significant ($p < 0.05$; McNemar’s) compared to re-ranking with the same discriminator on each dataset. We only observe such improvement with the oracle discriminator.
}
\vspace{-16pt}
\end{table*}

\section{LLM-Based Discriminators}
\label{intrin_eval}

%
While we have shown that iterative correction and tree search work well with oracle discriminators, it remains unclear whether LLM-based discriminators can correctly assess language agents’ actions (\textit{RQ2}).
To answer this question, we leverage generator outputs in the simulation experiments and re-label them with ground-truths to evaluate the LLMs' discrimination accuracy (Appendix \ref{app:intrin_eval_data}).
%

\subsection{Naive Discriminators}
\label{naive_llm}

%
As Table \ref{tab:intrin_eval} shows, most open-source LLMs have mediocre discrimination abilities.
To improve their accuracy \citep{arora2023learning, zhu-etal-2023-solving}, we further fine-tune the LLMs to classify the ground truth plans and incorrect ones sampled from the generator (Appendix \ref{app:discriminator_data}).
After fine-tuning, CodeLlama-13B-FT could reach the same level of performance as GPT-3.5.
In comparison, proprietary LLMs exhibit stronger discrimination abilities, with GPT-4 achieving the best performance across all three datasets.
Nonetheless, due to its high cost, we will use GPT-3.5 as the representative proprietary LLM in our experiments.
%

\subsection{Observation-Enhanced Discriminators}
\label{enhanced_llm}

%
To improve LLMs' discrimination abilities, we conduct an error analysis for CodeLlama-13B on its worst-performing intrinsic evaluation set, Bird.
We sample 50 pairs of SQL queries from the Bird intrinsic evaluation set with incorrect predictions.
In 25 of the 50 pairs (50\%), CodeLlama-13B assigns a higher score to non-executable SQL queries.
Consequently, no matter using which planning method, language agents could hardly perform well with such discriminators. 
Motivated by our error analysis, we first propose to add a program executability check as a safeguard for LLMs.
If a program is non-executable, our discriminator would discard LLMs' score and return 0.
Otherwise, it returns the original LLM score.
Besides executability check, we incorporate the execution results of predicted programs (first 5 table rows of SQL queries or answer of Python program) into the in-context examples and fine-tuning data \citep{pmlr-v202-ni23b-lever}.
If a program is non-executable, we use \texttt{ERROR} to represent its execution result. 
Evaluation results (Table \ref{tab:enhanced_llm}) show that these two non-oracle environmental observations can effectively improve LLMs' discrimination accuracy. 
Enhanced with environmental observations, CodeLlama-13B can obtain up to 25.4, 30.2, and 8.4 points absolute accuracy gain on Spider, Bird, and GSM8K, respectively.
For the other two models, we also observe significant gains compared to the naive discriminator baseline.
Such notable improvements also highlight the importance of filtering out non-executable programs, or invalid plans, during planning.
%

\section{End-to-End Evaluation}
\label{e2e}

%
While we have evaluated their discrimination abilities with a fixed test set, to answer \textit{RQ2}, we wonder if LLMs can correctly assess constantly changing sets of programs in actual planning processes.
To this end, we evaluate the end-to-end performance of language agents with LLM-based discriminators and the three planning methods. 
%

\subsection{Text-to-SQL Parsing}
\label{text2sql}

%
As shown in Table \ref{tab:text2sql}, agents using naive LLM-based discriminators do not perform well on text-to-SQL parsing.
On Spider, the re-ranking agent using CodeLlama-13B-FT has the best accuracy (61.5), which is still lower than greedy generation (62.3) that requires no planning and is more efficient.
On Bird, GPT-3.5-Turbo and re-ranking show an accuracy of 18.0, which is slightly higher than greedy generation (16.0).
In addition to the mediocre performance, we find that when using naive discriminators, iterative correction and tree search consistently show worse or the same performance as re-ranking.
These results mostly agree with our findings in previous experiments that
\textbf{(1)} advanced planning methods need strong discriminators, and
\textbf{(2)} naive LLM-based discriminators are not accurate enough.
After enhancing the discriminators with two environmental observations (Section \ref{enhanced_llm}), we effectively improve the agents' performance without any modifications to the generator or the planning methods.
In 5 of the 6 experiments, CodeLlama-13B-FT$^{E}$ results in the best execution accuracy among all discriminators.
It also leads to the overall best performance on Spider with re-ranking (70.3) and on Bird with iterative correction (26.3), showing the effectiveness of fine-tuning LLMs for discrimination and using environmental observations.
While the performance gains are satisfying, our implementation also takes the latency issue into consideration (Section \ref{oracle_results}).
For instance, on some BIRD databases with 10K or more rows, the runtime of a single SQL execution can take more than 10 minutes. 
To mitigate this issue, we speed up the executions by fetching only the first 5 rows. 
Additionally, we implement a 60-second timeout interruption for each SQL execution, which aligns to the limit in the official evaluation script.
It turns out that the environmental observations help to reduce the end-to-end latency (Table \ref{tab:bird_efficiency}), especially for tree search.
We think executability check is the main reason for this latency improvement. 
Since compilers and interpreters are heavily optimized, the check itself does not take much time. 
By quickly identifying incorrect programs, it helps to prune the search space, thus reducing the overall latency of re-ranking and tree search. 
For iterative correction, the latency is increased when using GPT-3.5-Turbo$^E$ and CodeLlama-13B-FT$^E$ as discriminators.
This is because these two discriminators are more accurate and allow iterative correction to run more planning loops, as stated in Section \ref{oracle_results} and Appendix \ref{app:ic_explanation}. 
%

\subsection{Mathematical Reasoning}
\label{gsm8k}

%
The most interesting result in mathematical reasoning evaluation (Table \ref{tab:gsm8k}) is the failure of iterative correction with naive discriminators.
When prompting the generator CodeLlama-13B for 0-shot correction, it would disregard the instruction to ``generate a fixed python program'' (Appendix \ref{app:prompts}), copy the program to be modified, and generate explanations and correction steps in natural language.
Such natural language steps, usually having some lexical overlap with the math problem, would increase the discrimination score of LLMs while being non-executable.
As a result, our iterative correction agent only has 10.2 answer accuracy when using CodeLlama-13B to evaluate its own generation.
While this issue also exists when using GPT-3.5-Turbo as the discriminator, it is less severe because GPT would sometimes assign a high score ($> 0.99$) to the initial Python program.
These scores trigger an early exit condition in iterative correction (Section \ref{planning_methods}) and stop the agent from calling the generator to add any natural language, thus avoiding the issue.
These findings echo related analysis on self-correction \citep{stechly2023gpt, valmeekam2023investigating, huang2024llm_selfcorrect}.
With an executability check, enhanced discriminators help mitigate this issue in iterative correction, which now achieves better performance (42.2 and 48.4) than greedy generation (39.4).
Overall, the tree search agent using GPT-3.5-Turbo$^E$ achieves the best answer accuracy. 
Nevertheless, McNemar’s test finds no difference ($p > 0.05$) between the performance of re-ranking (47.6) and that of iterative correction (48.4) or tree search (51.0).
%

\begin{table}[t]
\scriptsize
\centering
\begin{tabular}{
@{\hspace{0pt}}l@{\hspace{8pt}}ccc@{\hspace{0pt}}
}
\toprule
\textbf{Discriminators} & \textbf{Re-ranking} & \textbf{Iter. Correct.} & \textbf{Tree Search} \\
\midrule
CodeLlama-13B & 17.3 & 76.1 & 296.0 \\
GPT-3.5-Turbo & 24.4 & 41.0 & 405.2 \\
CodeLlama-13B-FT & 17.1 & 73.6 & 385.2 \\
\midrule
CodeLlama-13B$^{E}$ & 17.1 (-0.2) & 67.9 (-8.2) & 272.9 (-23.1) \\
GPT-3.5-Turbo$^{E}$ & 17.9 (-6.5) & 49.5 (+8.5) & 262.8 (-142.4) \\ 
CodeLlama-13B-FT$^{E}$ & 16.4 (-0.7) & 84.5 (+10.9) & 266.3 (-118.9) \\
\bottomrule
\end{tabular}
\vspace{-6pt}
\caption{\label{tab:bird_efficiency}
Average end-to-end inference time per example (seconds) on Bird. Notations have the same meaning as in Table \ref{tab:text2sql}.  For each observation-enhanced discriminator, we calculate the difference between its average inference time and that of its corresponding naive discriminator (base model).}
\vspace{-5.5pt}
\end{table}
\begin{table}[t]
\small
\centering
\begin{tabular}{
@{\hspace{0pt}}l@{\hspace{1.5pt}}
@{\hspace{1.5pt}}c@{\hspace{1.5pt}}
@{\hspace{1.5pt}}c@{\hspace{1.5pt}}
@{\hspace{1.5pt}}c@{\hspace{0pt}}
}
\toprule
\textbf{Discriminators} & \textbf{Re-ranking} & \textbf{Iter. Correct.} & \textbf{Tree Search}$^{\ddag}$ \\
\midrule
CodeLlama-13B & 39.7 & 10.2 & \textbf{41.0} \\
GPT-3.5-Turbo & 47.0 & 37.0 & \underline{\textbf{50.0}} \\
\midrule
CodeLlama-13B$^{E}$ & 42.8 & 42.2 & \textbf{46.0} \\
GPT-3.5-Turbo$^{E}$ & 47.6 & 48.4 & \underline{\textbf{51.0}} \\ 
\midrule
Oracle Simulation & \multirow{2}{*}{64.1} & \multirow{2}{*}{66.0} & \multirow{2}{*}{73.0} \\
($\tau = 1.0$) & & & \\
\bottomrule
\end{tabular}
\vspace{-6pt}
\caption{\label{tab:gsm8k}
End-to-end answer accuracy on GSM8K (Greedy Gen = 39.4). Notations have the same meaning as in Table \ref{tab:text2sql}. McNemar's does not find difference between methods on GSM8K. $^{\ddag}$Tree search is evaluated on 100 randomly selected examples from the 500 evaluation examples due to slow inference speed (Figure \ref{fig:gsm8k}). For McNemar's, we compare tree search results with those of re-ranking on the same 100 examples.
}
\vspace{-5.5pt}
\end{table}
\begin{table*}[t]
\small
\centering
\begin{tabular}{lcccccc}
\toprule
\multirow{2}{*}{\textbf{Error Type}} & \multicolumn{2}{c}{\textbf{Spider}} & \multicolumn{2}{c}{\textbf{Bird}} & \multicolumn{2}{c}{\textbf{GSM8K}} \\
\cmidrule(lr){2-3}\cmidrule(lr){4-5}\cmidrule(lr){6-7}
 & Iter. Correct. & Tree Search & Iter. Correct. & Tree Search  & Iter. Correct. & Tree Search \\
\midrule
Discrimination & 29 (78.4\%) & 17 (60.7\%) & 9 (52.9\%) & 12 (50.0\%) & 30 (62.5\%) & 6 (66.7\%) \\
Exploration & 8 (21.6\%) & 11 (39.3\%) & 8 (47.1\%) & 12 (50.0\%) & 18 (37.5\%) & 3 (33.3\%) \\
\midrule
Total & 37 & 28 & 17 & 24 & 48 & 9 \\
\bottomrule
\end{tabular}
\vspace{-6pt}
\caption{\label{tab:error_analysis}
Error analysis of examples where re-ranking outperforms advanced planning methods. We list the actual number of error cases and their percentages in parenthesis for each dataset and planning method.
}
\vspace{-16pt}
\end{table*}

\subsection{Analysis}
\label{e2e_analysis}

%
To better understand the end-to-end evaluation results, we conduct an in-depth analysis of examples where re-ranking returns the correct program, but iterative correction or tree search does not (Table \ref{tab:error_analysis}).
Specifically, we analyze cases of the strongest discriminators, CodeLlama-13B-FT$^{E}$ for text-to-SQL parsing and GPT-3.5-Turbo$^E$ for mathematical reasoning, and divide them into two kinds of errors.
\textbf{(1)} \textit{Discrimination error}: The discriminator assigns a higher score for wrong programs than correct ones, which is not recoverable by any planning method.
\textbf{(2)} \textit{Exploration error}: The planning method has not found the correct program before termination.
Our analysis suggests that:
\textbf{LLM-based discriminators have not yet met the needs of advanced planning methods.}
Across all datasets, 50\% or more discrimination errors are observed in each planning method.
On Spider, the number of such errors in iterative correction is as large as 29 out of 37 (78.4\%).
In fact, among the 29 errors, iterative correction has already found the correct SQL queries for 15 (40.5\% of the total 37 errors) of them.
However, not only does the discriminator fail to trigger early exits, but it also assigns a higher score for wrong SQL queries in new iterations.
Consequently, these erroneous SQL queries override the originally correct ones, leading to an overall performance drop.
The same issue is also serious in tree search.
When an incorrect partial program receives a high discrimination score, tree search will commit to it and hardly explore other possibilities, including the correct partial programs.
Such discrimination errors usually cannot be recovered by the planning methods themselves, unless they find another correct program with even higher scores.
This finding also demonstrates that determining early exits using oracle information in iterative correction may introduce a larger benefit than previously thought \citep{huang2024llm_selfcorrect}.
\textbf{Advanced planning methods need more thorough exploration.}
For the remaining cases, we observe that advanced planning methods have not found a correct program before terminating, which we call exploration errors.
This kind of error circles our discussion back to the accuracy-efficiency trade-off mentioned in our simulation experiments with oracle (Section \ref{oracle_results}).
Indeed, we can extend the exploration of planning methods in various ways, such as loosening termination conditions, increasing the number of generation samples for each step, and adjusting some hyperparameters for more diverse program samples.
Yet, all these adjustments can slow down the planning methods and reduce the language agents' efficiency.
Additionally, we note that these strategies may not always result in better performance, as the discriminators may give unseen wrong programs a higher score.
For these reasons, \textbf{iterative correction and tree search cannot gain decent improvement} over re-ranking with the same LLM-based discriminator.
On text-to-SQL parsing, tree search even shows worse performance than re-ranking when using CodeLlama-13B-FT$^{E}$ (Table \ref{tab:text2sql}: 67.5 vs 70.3 on Spider; 21.7 vs 23.7 on Bird).
More surprisingly, on GSM8K, advanced planning methods may not perform much better than re-ranking even with the oracle discriminator ($p > 0.05$; McNemar’s).
Admittedly, some of the performance gains appear considerable, but McNemar’s tells us there are still decent chances of the simpler agent outperforming a more complex one (Appendix \ref{app:stat_test}).
%

\section{Conclusions}
\label{conclusion}

%
This paper presents a thorough investigation into the relationship between discrimination accuracy and performance of planning methods in language agents.
Through comprehensive experiments on text-to-SQL parsing and mathematical reasoning, we find that: 
Discrimination accuracy strongly correlates with the overall performance of language agents using different planning methods and also affects their efficiency (\textit{answer to RQ1}). 
LLM-based discriminators can correctly assess a decent number of language agents' actions with their environmental observations, but they are still not accurate enough for advanced planning methods (\textit{answer to RQ2}).
Future research should investigate the development of more accurate discrimination models for language agents, e.g.\ by improving their grounded understanding of execution results beyond error signals.
%

\section*{Limitations}

%
\noindent \textbf{Experiments with Other Models.} 
In this study, we focus on studying the generation and discrimination of instruction-tuned LLMs that have seen code data during pre-training.
This consideration is because: 
\textbf{(a)} They may have better in-context learning performance on our two tasks, text-to-SQL parsing and mathematical reasoning with program-of-thought \citep{ni2023l2ceval}; 
\textbf{(b)} We want to leverage their 0-shot instruction following capabilities in iterative correction for fair comparisons with other planning methods;
\textbf{(c)} For GSM8K problems, LLMs tend to generate natural language plans instead of programs with 2-shot prompting, and some instructions other than in-context examples help to mitigate this issue.
Future research may extend our study to other LLMs of code and conduct an ablation study of instruction-tuning's impact on models' discrimination accuracy.
\noindent \textbf{Experiments with Natural Language Plans.} 
Our study focuses on the generation and discrimination of formal language plans, i.e., programs, as they can directly interact with the environment. 
Although feasible for mathematical reasoning \citep{wei2022cot}, natural language plans require another semantic parsing step to convert them into actions defined in the corresponding environment, which may introduce intermediate errors and add noise to our analysis.
Therefore, we conduct the experiments with formal language plans using LLMs trained on code data.
As a future direction, it would be interesting to extend our study to natural language plans and see how the intermediate semantic parsing step would affect the overall performance of agents for mathematical reasoning.
\noindent \textbf{Impact of Generators on Planning Methods.}
While our work focuses on studying the relationship between different discriminators and planning methods, we acknowledge that the generator can also actively affect different planning methods.
For example, we can transform the generator's perplexity into a probability and multiply it by the discriminator's score.
We exclude such uses of the generator because in our preliminary experiments, we find that incorporating its perplexity leads to mixed results.
These results make it even harder to analyze how language agents behave when using different planning methods. Thus, we exclude the generator to have a clear picture of how discriminators can affect planning methods. Nevertheless, it is worth studying the generator's impact on planning methods in future work.

\section*{Acknowledgements}
We would like to thank colleagues from the OSU NLP group for their thoughtful comments. 
This research was supported in part by a sponsored award from Cisco Research, NSF IIS-1815674, NSF CAREER \#1942980, NSF OAC-2112606, and Ohio Supercomputer Center \citep{OhioSupercomputerCenter1987}. 
The views and conclusions contained herein are those of the authors and should not be interpreted as representing the official policies, either expressed or implied, of the U.S. government.
The U.S. Government is authorized to reproduce and distribute reprints for Government purposes notwithstanding any copyright notice herein.

\bibliography{custom}

\appendix


\setcounter{table}{0}
\renewcommand\thetable{\Alph{section}.\arabic{table}}
\setcounter{figure}{0}
\renewcommand\thefigure{\Alph{section}.\arabic{figure}}

\section{More Implementation Details}
\label{app:implementation}

\subsection{Text-to-SQL Parsing Evaluation Sets}
\label{app:text2sql_data}

%
For text-to-SQL parsing, we sub-sample the development splits of each dataset, Spider and Bird, following three steps:
\textbf{(1)} categorize development set examples by difficulty levels defined in each dataset,
\textbf{(2)} randomly select a database and choose one associated example, and
\textbf{(3)} repeat step 2 until we have 100 samples for each difficulty level.
In this way, we ensure a uniform distribution across different difficulty levels and database.
Since there are four and three difficulty levels in Spider and Bird, respectively, our evaluation sets have 400 and 300 examples for each dataset.
Text-to-SQL parsing models, including LLMs, may show lower performance on our evaluation sets because of their uniformly distributed difficulty (100 examples per level).
In comparison, the original datasets have skewed distributions towards easier examples.
Spider's development set has 248 (24.0\%) examples at easy level and 446 (43.1\%) examples at medium level, while the hard and extra hard examples only sum up to 32.9 \% of the 1,034 examples.
In Bird, 925 out of the 1,534 (60.3\%) development set examples are at simple level, 465 examples (30.3\%) are at moderate level, and only 144 examples (9.4\%) are at challenging level.
Our evaluation sets normalize these skewed distributions and make the macro averages of model performance less biased (Section \ref{tasks_datasets}).
%

\subsection{Intrinsic Evaluation Data}
\label{app:intrin_eval_data}

%
\noindent To evaluate LLMs' discrimination performance, we reuse the generation results from our oracle-simulation experiments (Section \ref{intrin_eval}).
Specifically, we use the evaluation scripts to re-label the generated programs in simulated re-ranking experiments (accuracy threshold $\tau = 1.0$).
Then, we construct our intrinsic evaluation sets based on the relabeled programs (Table \ref{tab:intrin_eval_data_app}).
Intuitively, the number of program batches for each dataset is the same as the end-to-end evaluation examples we have, and the the number of programs is all unique programs we can get from the batches.
To pair the programs and calculate discrimination accuracy, we iterate through each batch and enumerate combinations of correct and wrong programs within the batch.
We do not include cross-batch pairs, as those do not align with our end-to-end evaluation settings.
For discrimination accuracy, we enumerate pairs of correct and wrong programs and ask LLMs to select the better one.
For classification F1, we let LLMs predict the correctness of each individual program.
For Hit@1 and MRR, we use LLMs to score the batches of programs in simulation experiments.

\begin{table}[t]
\small
\centering
\begin{tabular}{
@{\hspace{0pt}}l@{\hspace{1.5pt}}
@{\hspace{1.5pt}}c@{\hspace{3pt}}
@{\hspace{3pt}}c@{\hspace{3pt}}
@{\hspace{3pt}}c@{\hspace{0pt}}
}
\toprule
 & \textbf{Spider} & \textbf{Bird} & \textbf{GSM8K} \\
\midrule
Number of Programs & 1,221 & 1,291 & 2,453 \\
Number of Program Pairs & 409 & 269 & 1,238 \\
Number of Program Batches & 400 & 300 & 500 \\
\bottomrule
\end{tabular}
\caption{\label{tab:intrin_eval_data_app}
Statistics of our intrinsic evaluation sets.
}
\vspace{-11pt}
\end{table}

\subsection{Data for Discriminator LMs}
\label{app:discriminator_data}

%
\noindent For text-to-SQL parsing, we perform 2-fold cross-validation on the training sets to synthesize incorrect SQL queries for each example \citep{chen-etal-2023-error}.
We prompt the LM using one pair of correct and wrong SQL queries (labeled with ``Yes'' and ``No''), also retrieved by BM25 (Section \ref{discriminator}).
Alternatively, we fine-tune the LM on the entire training set with ground-truth and synthesized SQL queries to generate ``Yes'' or ``No.''
For mathematical reasoning, we annotate two incorrect python programs for the two examples used in generator.
Similar to text-to-SQL parsing, we use the two program pairs to prompt LMs for binary question answering.
Since the training set of GSM8K is not annotated with program of thoughts, we are not able to fine-tune LMs on this dataset.
%

\subsection{Implemendation of Oracle Discriminator}
\label{app:oracle_implementation}

%
For text-to-SQL parsing, our oracle uses the first five rows in execution results of the predicted and gold SQL query and calculate the table cell overlap.
More specifically, the calculation is similar to span F1 in machine reading comprehension.
Our oracle function first compares each row in the execution results head-to-head under a strong assumption that the rows are ordered.
Although strict, this assumption is helpful for evaluating the correctness of SQL queries with an \texttt{ORDER BY} clause.
Then, the function count how many table cells overlap with each other in an unordered manner.
We divide the number of overlapping cells by the total number of cells in execution results of the gold SQL query (precision) and the predicted one (recall).
Finally, we compute the harmonic mean of these two numbers to get the oracle score (F1).
For instance, given ``-{}- countryid: 1, 2, 4, 5 -{}- countryname: usa, germany, japan, italy'' as the gold execution result and ``-{}- countryid: 1, 4, 6 -{}- countryname: usa, japan, japan'' as the result of predicted SQL query.
We compare (1, usa), (4, japan), and (6, japan) the first, second, and third row in the gold result, respectively.
They have 2, 0, and 1 overlapping table cells, respectively.
Thus, we have our precision to be $3/8 = 0.375$ and recall to be $3/6 = 0.5$.
The oracle's score would be:
\[\frac{2 \cdot 0.375 \cdot 0.5}{0.375 + 0.5} = 0.43\]
\noindent For mathematical reasoning, our oracle directly checks if the predicted answer equals to the ground-truth. 
If the answer is \texttt{None} (non-executable program) or does not equal to the ground-truth, it returns 0.
Otherwise, it returns 1.
%

\newpage
\setcounter{table}{0}
\renewcommand\thetable{\Alph{section}.\arabic{table}}
\setcounter{figure}{0}
\renewcommand\thefigure{\Alph{section}.\arabic{figure}}

\section{McNemar’s Test for Statistical Significance}
\label{app:stat_test}

%
\noindent We measure the statistical significance of performance gains using the exact McNemar’s Test\footnote{\url{https://www.statsmodels.org/dev/generated/statsmodels.stats.contingency_tables.mcnemar.html}} \citep{McNemar}.
We choose the test's exact binomial version because our sample sizes are relatively small \citep{ExactBinomialTest}, and the first two significant digits of $p$-values are the same for this binomial version and the original chi-square version in our tests. 
Intuitively, this test measures how likely the weaker method can still outperform the stronger one.
For example, we consider the comparison between tree search and iterative correction on Bird when using CodeLlama-13B-FT$^{E}$ as the discriminator (Section \ref{oracle_results}).
By computing a $2\times2$ contingency table (Table \ref{tab:cont_table}), McNemar’s Test focuses on the 40 examples where only one of the two method have predicted correctly.
Specifically, there are 25 examples that iterative correction finds the correct answer, but tree search does not, which is the source of performance gain.
Also, there are 15 examples that iterative correction fails, but tree search succeeds.
According to McNemar’s Test, these 15 (37.5\% of the total 40) examples result in a $p$-value of 0.15, meaning there is still some chance for tree search to outperform iterative correction.
In contrast, suppose there are only 10 examples that iterative correction finds the correct answer, but tree search does not.
Meanwhile, there are no examples that iterative correction fails, but tree search succeeds.
Then, we can still observe the same number of accuracy gain, but it is now statistically different because it is almost impossible for tree search to outperform iterative correction (0 out of 10).
The same rationale also applies to the results of other tests in Section \ref{e2e}.

\begin{table}[t]
\small
\centering
\begin{tabular}{lcc}
\toprule
 & IC Correct & IC Wrong \\
\midrule
TS Correct & 73 & 15 \\
TS Wrong & 25 & 187 \\
\bottomrule
\end{tabular}
\caption{\label{tab:cont_table}
Contingency table for tree search (TS) and iterative correction (IC) on Bird, using CodeLlama-13B-FT$^{E}$ as the discriminator (Section \ref{oracle_results}).
}
\vspace{-11pt}
\end{table}



\setcounter{table}{0}
\renewcommand\thetable{\Alph{section}.\arabic{table}}
\setcounter{figure}{0}
\renewcommand\thefigure{\Alph{section}.\arabic{figure}}

\section{More Analysis on Simulation Experiments with Oracle}
\label{app:ic_explanation}

%
In Section \ref{oracle_results}, we conclude that ``the inference time for iterative correction increases as the accuracy threshold is raised, suggesting more iterations are required to derive a correct answer.'' Though counter-intuitive, we verify that this conclusion is correct because an accurate discriminator would avoid triggering early exit by mistake in iterative correction.
To give a concrete example, we compare the number of iterations needed on Spider when using oracle-based discriminator with accuracy threshold 0.6 and 1.0. When the threshold is 0.6, the planning method takes 1.275 iterations on average. When the threshold is 1.0, the planning method takes 1.725 iterations on average, 1.35 times more than the previous case. This roughly aligns with the increase in average inference time: When the threshold is 0.6, iterative correction takes an average of 21.5 seconds per example. When the threshold is 1.0, the average inference time is 27.9, which is 1.29 times more.
This counter-intuitive observation is mainly caused by the early exit of iterative correction methods. In other words, if the discrimination score is high enough (>0.99) or is not improved for 3 consecutive iterations, the planning method would stop making more corrections (Lines 214--217). When the discriminator is not accurate, it would (a) give high scores to wrong plans or (b) keep assigning the same or even lower scores to improved plans. As a result, the planning method would stop quickly without finding a better plan. On the other hand, an accurate discriminator would predict less extreme scores and increase them for even small improvements in the plan, thus allowing the method to iterate 3 more times. This results in the increased inference time per example.
%

\newpage
\onecolumn

\setcounter{table}{0}
\renewcommand\thetable{\Alph{section}.\arabic{table}}
\setcounter{figure}{0}
\renewcommand\thefigure{\Alph{section}.\arabic{figure}}

\section{Prompt Examples}
\label{app:prompts}

\begin{table*}[h]
\small
\centering
\begin{tabular}{p{0.9\linewidth}}
\toprule
\texttt{Given database schema and a question in natural language, generate the corresponding SQL query.\newline
\newline
-{}- Database climbing: \newline
-{}- Table mountain: mountain\_id, name, height, prominence, range, country \newline
-{}- Table climber: climber\_id, name, country, time, points, mountain\_id \newline
-{}- Question: How many distinct countries are the climbers from? \newline
-{}- SQL: \newline
SELECT COUNT(DISTINCT country) FROM climber; \newline
\newline
-{}- Database concert\_singer: \newline
-{}- Table stadium: stadium\_id, location, name, capacity, highest, lowest, average \newline
-{}- Table singer: singer\_id, name, country, song\_name, song\_release\_year, age, is\_male \newline
-{}- Table concert: concert\_id, concert\_name, theme, stadium\_id, year \newline
-{}- Table singer\_in\_concert: concert\_id, singer\_id \newline
-{}- Question: What are all distinct countries where singers above age 20 are from? \newline
-{}- SQL: \newline
SELECT
} \\
\bottomrule
\end{tabular}
\caption{\label{tab:gen_prompt_text2sql}
An example prompt for 1-shot generation (text-to-SQL parsing).
}
\end{table*}

\begin{table*}[h]
\small
\centering
\begin{tabular}{p{0.9\linewidth}}
\toprule
\texttt{Given database schema and a question in natural language, correct the buggy SQL query and generate a fixed SQL query.\newline
\newline
-{}- Database concert\_singer: \newline
-{}- Table stadium: stadium\_id, location, name, capacity, highest, lowest, average \newline
-{}- Table singer: singer\_id, name, country, song\_name, song\_release\_year, age, is\_male \newline
-{}- Table concert: concert\_id, concert\_name, theme, stadium\_id, year \newline
-{}- Table singer\_in\_concert: concert\_id, singer\_id \newline
-{}- Question: What are all distinct countries where singers above age 20 are from? \newline
-{}- Buggy SQL: \newline
SELECT DISTINCT country FROM singer WHERE age > 20; \newline
-{}- Fixed SQL: \newline
SELECT
} \\
\bottomrule
\end{tabular}
\caption{\label{tab:gen_prompt_text2sql_ic}
An example prompt for 0-shot iterative correction (text-to-SQL parsing).
}
\end{table*}

\newpage

\begin{table*}[h]
\small
\centering
\begin{tabular}{p{0.9\linewidth}}
\toprule
\texttt{Answer the following Yes/No question: Is the SQL correct given the utterance? \newline
\newline
-{}- Utterance: How many different countries are all the swimmers from? \newline
-{}- SQL: \newline
SELECT COUNT(DISTINCT nationality) FROM swimmer; \newline
-{}- Answer: Yes \newline
\newline
-{}- Utterance: How many different countries are all the swimmers from? \newline
-{}- SQL: \newline
SELECT DISTINCT country FROM swimmer; \newline
-{}- Answer: No \newline
\newline
-{}- Utterance: What are all distinct countries where singers above age 20 are from? \newline
-{}- SQL: \newline
SELECT DISTINCT country FROM singer WHERE age > 20; \newline
-{}- Answer:
} \\
\bottomrule
\end{tabular}
\caption{\label{tab:eval_prompt_text2sql_base}
An example prompt for 1-shot discrimination (text-to-SQL parsing). For discrimination, each in-context example has a pair of correct and wrong programs.
}
\end{table*}

\begin{table*}[h]
\small
\centering
\begin{tabular}{p{0.9\linewidth}}
\toprule
\texttt{Answer the following Yes/No question: Is the SQL correct given the utterance and its result? \newline
\newline
-{}- Utterance: How many different countries are all the swimmers from? \newline
-{}- SQL: \newline
SELECT COUNT(DISTINCT nationality) FROM swimmer; \newline
-{}- Result: \newline
-{}- count(distinct nationality): 7 \newline
-{}- Answer: Yes \newline
\newline
-{}- Utterance: How many different countries are all the swimmers from? \newline
-{}- SQL: \newline
SELECT DISTINCT country FROM swimmer; \newline
-{}- Result: \newline
ERROR \newline
-{}- Answer: No \newline
\newline
-{}- Utterance: What are all distinct countries where singers above age 20 are from? \newline
-{}- SQL: \newline
SELECT DISTINCT country FROM singer WHERE age > 20; \newline
-{}- Result: \newline
-{}- country: Netherlands, United States, France \newline
-{}- Answer:
} \\
\bottomrule
\end{tabular}
\caption{\label{tab:eval_prompt_text2sql_pro}
An example prompt for 1-shot discrimination with execution results (text-to-SQL parsing). For discrimination, each in-context example has a pair of correct and wrong programs.
}
\end{table*}

\newpage

\begin{table*}[h]
\small
\centering
\begin{tabular}{p{0.9\linewidth}}
\toprule
\texttt{\#\# Given questions in the comment, use python programs to produce the correct answers with the 'answer' variable. \newline
\newline
\#\# James takes 2 Tylenol tablets that are 375 mg each, every 6 hours. How many mg does he take a day? \newline
\#\# Python Program: \newline
mg\_tylenol\_per\_tablet = 375 \newline
mg\_tylenol\_taken\_each\_time = 2 * mg\_tylenol\_per\_tablet \newline
hours\_per\_day = 24 \newline
times\_per\_day = hours\_per\_day / 6 \newline
mg\_each\_day = mg\_tylenol\_taken\_each\_time * times\_per\_day \newline
answer = mg\_each\_day \newline
\newline
\#\# There were 63 Easter eggs in the yard. Hannah found twice as many as Helen. How many Easter eggs did Hannah find? \newline
\#\# Python Program: \newline
n\_easter\_eggs = 63 \newline
unit\_times = 2 \newline
total\_units = unit\_times + 1 \newline
n\_easter\_eggs\_per\_unit = n\_easter\_eggs / total\_units \newline
n\_easter\_eggs\_helen = n\_easter\_eggs\_per\_unit * 1 \newline
n\_easter\_eggs\_hannah = n\_easter\_eggs\_per\_unit * 2 \newline
answer = n\_easter\_eggs\_hannah \newline
\newline
\#\# Gloria is shoe shopping when she comes across a pair of boots that fit her shoe budget. However, she has to choose between the boots and two pairs of high heels that together cost five dollars less than the boots. If one pair of heels costs \$33 and the other costs twice as much, how many dollars are the boots? \newline
\#\# Python Program:
} \\
\bottomrule
\end{tabular}
\caption{\label{tab:gen_prompt_gsm8k}
An example prompt for 2-shot generation (mathematical reasoning).
}
\end{table*}

\begin{table*}[h]
\small
\centering
\begin{tabular}{p{0.9\linewidth}}
\toprule
\texttt{\#\# Given the question in the comment, correct the buggy python program and generate a fixed python program to produce the correct answer with the 'answer' variable. \newline
\newline
\#\# Gloria is shoe shopping when she comes across a pair of boots that fit her shoe budget. However, she has to choose between the boots and two pairs of high heels that together cost five dollars less than the boots. If one pair of heels costs \$33 and the other costs twice as much, how many dollars are the boots? \newline
\#\# Buggy Python Program: \newline
price\_boots = 50 \newline
price\_heels = 33 \newline
price\_heels\_twice = 2 * price\_heels \newline
price\_heels\_total = price\_heels + price\_heels\_twice \newline
price\_boots\_difference = price\_boots - price\_heels\_total \newline
answer = price\_boots\_difference \newline
\#\# Fixed Python Program:
} \\
\bottomrule
\end{tabular}
\caption{\label{tab:gen_prompt_gsm8k_ic}
An example prompt for 0-shot iterative correction (mathematical reasoning).
}
\end{table*}

\newpage

\begin{table*}[h]
\small
\centering
\begin{tabular}{p{0.9\linewidth}}
\toprule
\texttt{\#\# Answer the following Yes/No question: Is the python program correct given the problem in the comment? \newline
\newline
\#\# James takes 2 Tylenol tablets that are 375 mg each, every 6 hours. How many mg does he take a day? \newline
\#\# Python Program: \newline
mg\_tylenol\_per\_tablet = 375 \newline
mg\_tylenol\_taken\_each\_time = 2 * mg\_tylenol\_per\_tablet \newline
hours\_per\_day = 24 \newline
times\_per\_day = hours\_per\_day / 6 \newline
mg\_each\_day = mg\_tylenol\_taken\_each\_time * times\_per\_day \newline
answer = mg\_each\_day \newline
\#\# Answer: Yes \newline
\newline
\#\# James takes 2 Tylenol tablets that are 375 mg each, every 6 hours. How many mg does he take a day? \newline
\#\# Python Program: \newline
mg\_per\_tablet = 375 \newline
n\_tablets\_per\_day = 2 \newline
n\_tablets\_per\_6hrs = n\_tablets\_per\_day / 6 \newline
mg\_per\_6hrs = mg\_per\_tablet * n\_tablets\_per\_6hrs \newline
answer = mg\_per\_6hrs \newline
\#\# Answer: No \newline
\newline
\#\# There were 63 Easter eggs in the yard. Hannah found twice as many as Helen. How many Easter eggs did Hannah find? \newline
\#\# Python Program: \newline
n\_easter\_eggs = 63 \newline
unit\_times = 2 \newline
total\_units = unit\_times + 1 \newline
n\_easter\_eggs\_per\_unit = n\_easter\_eggs / total\_units \newline
n\_easter\_eggs\_helen = n\_easter\_eggs\_per\_unit * 1 \newline
n\_easter\_eggs\_hannah = n\_easter\_eggs\_per\_unit * 2 \newline
answer = n\_easter\_eggs\_hannah \newline
\#\# Answer: Yes \newline
\newline
\#\# There were 63 Easter eggs in the yard. Hannah found twice as many as Helen. How many Easter eggs did Hannah find? \newline
\#\# Python Program: \newline
eggs\_in\_yard = 63 \newline
eggs\_found\_by\_hannah = 2 * eggs\_in\_yard \newline
eggs\_found\_by\_helen = eggs\_found\_by\_hannah / 2 \newline
answer = eggs\_found\_by\_hannah \newline
\#\# Answer: No \newline
\newline
\#\# Gloria is shoe shopping when she comes across a pair of boots that fit her shoe budget. However, she has to choose between the boots and two pairs of high heels that together cost five dollars less than the boots. If one pair of heels costs \$33 and the other costs twice as much, how many dollars are the boots? \newline
\#\# Python Program: \newline
price\_boots = 50 \newline
price\_heels = 33 \newline
price\_heels\_twice = 2 * price\_heels \newline
price\_heels\_total = price\_heels + price\_heels\_twice \newline
price\_boots\_difference = price\_boots - price\_heels\_total \newline
answer = price\_boots\_difference \newline
\#\# Answer:
} \\
\bottomrule
\end{tabular}
\caption{\label{tab:eval_prompt_gsm8k_base}
An example prompt for 2-shot discrimination (mathematical reasoning). For discrimination, each in-context example has a pair of correct and wrong programs.
}
\end{table*}

\newpage

\begin{table*}[h]
\small
\centering
\begin{tabular}{p{0.9\linewidth}}
\toprule
\texttt{\#\# Answer the following Yes/No question: Is the python program correct given its result and the problem in the comment? \newline
\newline
\#\# James takes 2 Tylenol tablets that are 375 mg each, every 6 hours. How many mg does he take a day? \newline
\#\# Python Program: \newline
mg\_tylenol\_per\_tablet = 375 \newline
mg\_tylenol\_taken\_each\_time = 2 * mg\_tylenol\_per\_tablet \newline
hours\_per\_day = 24 \newline
times\_per\_day = hours\_per\_day / 6 \newline
mg\_each\_day = mg\_tylenol\_taken\_each\_time * times\_per\_day \newline
answer = mg\_each\_day \newline
\#\# Result: 3000.0 \newline
\#\# Answer: Yes \newline
\newline
\#\# James takes 2 Tylenol tablets that are 375 mg each, every 6 hours. How many mg does he take a day? \newline
\#\# Python Program: \newline
mg\_per\_tablet = 375 \newline
n\_tablets\_per\_day = 2 \newline
n\_tablets\_per\_6hrs = n\_tablets\_per\_day / 6 \newline
mg\_per\_6hrs = mg\_per\_tablet * n\_tablets\_per\_6hrs \newline
answer = mg\_per\_6hrs \newline
\#\# Result: 125.0 \newline
\#\# Answer: No \newline
\newline
\#\# There were 63 Easter eggs in the yard. Hannah found twice as many as Helen. How many Easter eggs did Hannah find? \newline
\#\# Python Program: \newline
n\_easter\_eggs = 63 \newline
unit\_times = 2 \newline
total\_units = unit\_times + 1 \newline
n\_easter\_eggs\_per\_unit = n\_easter\_eggs / total\_units \newline
n\_easter\_eggs\_helen = n\_easter\_eggs\_per\_unit * 1 \newline
n\_easter\_eggs\_hannah = n\_easter\_eggs\_per\_unit * 2 \newline
answer = n\_easter\_eggs\_hannah \newline
\#\# Result: 42 \newline
\#\# Answer: Yes \newline
\newline
\#\# There were 63 Easter eggs in the yard. Hannah found twice as many as Helen. How many Easter eggs did Hannah find? \newline
\#\# Python Program: \newline
eggs\_in\_yard = 63 \newline
eggs\_found\_by\_hannah = 2 * eggs\_in\_yard \newline
eggs\_found\_by\_helen = eggs\_found\_by\_hannah / 2 \newline
answer = eggs\_found\_by\_hannah \newline
\#\# Result: 126 \newline
\#\# Answer: No \newline
\newline
\#\# Gloria is shoe shopping when she comes across a pair of boots that fit her shoe budget. However, she has to choose between the boots and two pairs of high heels that together cost five dollars less than the boots. If one pair of heels costs \$33 and the other costs twice as much, how many dollars are the boots? \newline
\#\# Python Program: \newline
price\_boots = 50 \newline
price\_heels = 33 \newline
price\_heels\_twice = 2 * price\_heels \newline
price\_heels\_total = price\_heels + price\_heels\_twice \newline
price\_boots\_difference = price\_boots - price\_heels\_total \newline
answer = price\_boots\_difference \newline
\#\# Result: -49 \newline
\#\# Answer:
} \\
\bottomrule
\end{tabular}
\caption{\label{tab:eval_prompt_gsm8k_pro}
An example prompt for 2-shot discrimination with execution results (mathematical reasoning). For discrimination, each in-context example has a pair of correct and wrong programs.
}
\end{table*}

\end{document}